\documentclass[11pt]{article}

\usepackage[final]{acl}
\usepackage{amsmath}
\usepackage{times}
\usepackage{latexsym}
\usepackage{tcolorbox}
\tcbuselibrary{breakable}
\usepackage[T1]{fontenc}

\usepackage[utf8]{inputenc}

\usepackage{microtype}

\usepackage{inconsolata}

\usepackage{graphicx}
\usepackage[framemethod=tikz]{mdframed}

\usepackage[table]{xcolor}
\usepackage{booktabs}
\usepackage{amssymb}
\usepackage{algorithm}
\usepackage{algpseudocode}
\usepackage{float}
\newcommand{\meanstd}[2]{#1{\scriptsize$\pm$#2}}
\newcommand{\meanstdu}[2]{\underline{#1{\scriptsize$\pm$#2}}}
\newcommand{\meanstdb}[2]{\textbf{#1}{\scriptsize$\boldsymbol{\pm}\mathbf{#2}$}}

\title{PRISM Edit: One Vector for All Temporal Answers}

\author{
  \textbf{Chen Huang}\textsuperscript{1,\dag},
  \textbf{Qi Zheng}\textsuperscript{1,\dag},
  \textbf{Ruiqin Zheng}\textsuperscript{2},
  \textbf{Long Zeng}\textsuperscript{1,*},
  \textbf{Yuantong Xu}\textsuperscript{2,*}
\\
\\
  \textsuperscript{1}Tsinghua University,
  \textsuperscript{2}ByteDance
\\
  \small{
    \textsuperscript{\dag}Equal contribution.\quad
    \textsuperscript{*}Corresponding authors.
  }
\\
  \small{
    \textbf{Correspondence:} \href{mailto:zenglong@sz.tsinghua.edu.cn}{zenglong@sz.tsinghua.edu.cn}
  }
}

\begin{document}
\maketitle

\begin{abstract}
Model editing keeps large language models (LLMs) up to date without retraining, but temporal facts expose a limitation of the prevailing locate-and-edit paradigm: an update is not always a replacement. When a fact changes, the new answer should become current while the old answer may remain correct in historical time contexts. Building on this insight, we use causal tracing to show that LLMs already support this distinction via a two-stage internal computation: early MLP layers retrieve a time-agnostic subject representation, and later layers modulate it with temporal context to yield the time-correct answer. 
Motivated by this finding, we introduce \textbf{PRISM Edit}, which optimizes a single polysemous representation across temporal contexts and leverages the model's inherent modulation pathway to route it to temporally correct predictions without requiring any architectural modification. We evaluate on \textsc{TimeConflict}, a newly introduced temporal editing benchmark, and on temporally augmented \textsc{CounterFact}. PRISM Edit improves multiple core metrics over the best baseline, most notably +23.3 Temporal Consistency (TC) and +33.7 Current Relative-time Score (CRS) on LLaMA-3, while being more than $2\times$ faster. Code and data are publicly available at \url{https://github.com/CheerCHuang/PRISM-Edit}. 
\end{abstract}

\section{Introduction}
\label{sec:intro}

\begin{figure*}[t]
\centering
\includegraphics[width=\textwidth]{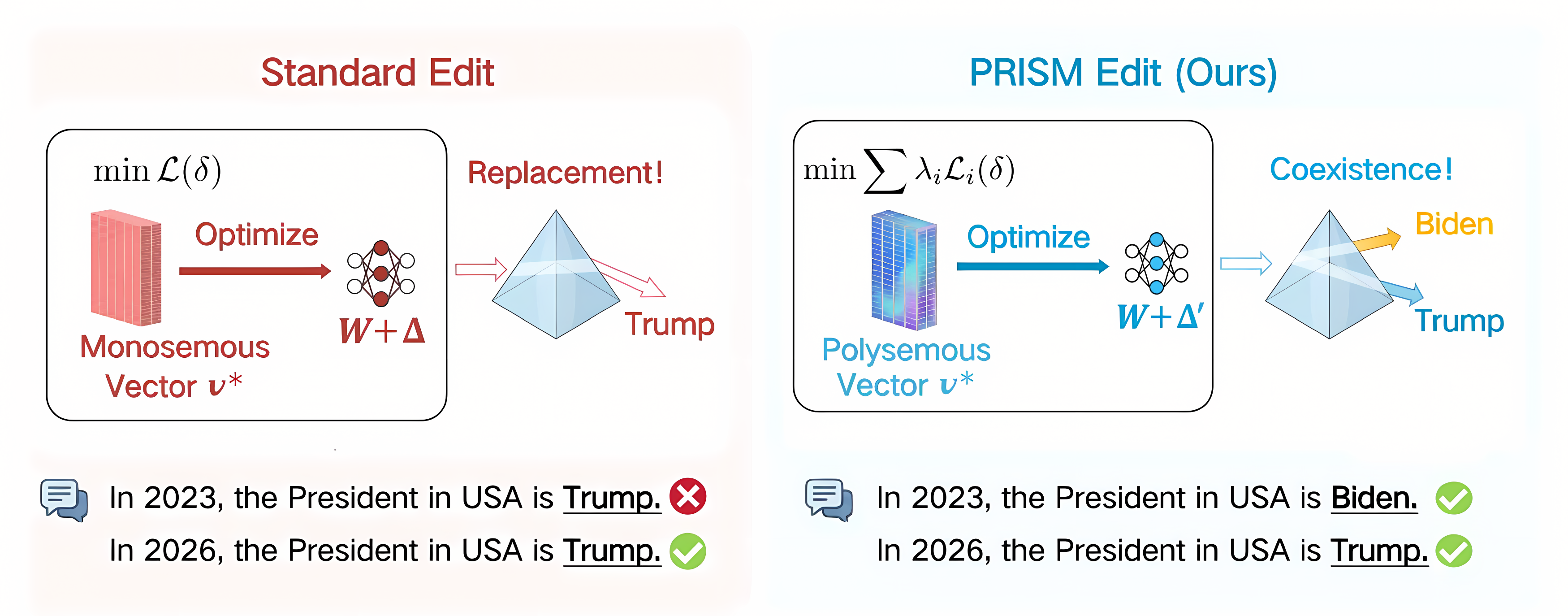}
\caption{\textbf{Standard editing vs.\ PRISM Edit.} Left: conventional locate-and-edit methods overwrite the factual association with a single new target, collapsing historical recall (``In 2023\ldots'' incorrectly returns \emph{Trump}). Right: PRISM Edit optimizes a single polysemous vector $v^*$ that encodes both temporal answers; the model's own downstream modulation routes it to the temporally correct prediction under each time context.}
\label{fig:the mainfig}
\end{figure*}

Large language models store extensive factual knowledge, yet the world they describe keeps changing. Knowledge editing allows models to continuously acquire new knowledge without full retraining, typically by modifying a small set of parameters associated with a target memory~\cite{decao2021editing,mitchell2022fast,meng2022locating,meng2023memit,zhang2024comprehensivestudyknowledgeediting,fang2025alphaedit}. 
However, existing methods implicitly assume that every edit is a wholesale replacement of the old fact. This assumption breaks down for \emph{temporal} facts, where the old
answer does not become wrong---it merely ceases to be current. For example, when the U.S. presidency transfers from \emph{Biden} to \emph{Trump} in January 2025, the new fact should become current, yet ``Who was the U.S.\ president in 2023?'' must still return \emph{Biden}. A temporal editor must therefore perform a harder operation: incorporate the new answer while preserving the historical conditions under which the old answer is still true. 

The difficulty lies in the locate-and-edit paradigm itself: it locates a factual association $(s, r, o)$ and overwrites it with a new target $o'$. This works for single-valued facts, but it mismatches temporal knowledge, where the same $(s, r)$ pair can map to different objects under different time contexts. As a result, standard editors often collapse time-conditioned behavior into a single edited answer, degrading historical recall after temporal edits~\cite{yin2024history,cheng2024multi,zhao2026multi} (Figure~\ref{fig:the mainfig}). Existing temporal-editing strategies attempt to patch this gap by storing separate parameters per time period~\cite{zhao2026multi} or splitting updates into multiple independent edits~\cite{yin2024history}. Yet these approaches leave a key question unanswered: if the unedited model can already answer time-conditioned prompts, why should editing ignore the model's own temporal computation?

We take a fundamentally different approach. Rather than imposing external temporal structure, we first ask \emph{how} the model internally resolves temporal ambiguity, and then design an editing method that works \emph{with}---not against---this mechanism. Through causal tracing on $(s, r, o, t)$ tuples (\S\ref{sec:trace}), we discover that subject retrieval at early MLP layers is time-agnostic, while temporal disambiguation emerges via later-layer modulation of the same subject signal. This finding motivates \textbf{PRISM Edit} (\textbf{P}olysemous \textbf{R}epresentation via \textbf{I}ntrinsic \textbf{S}ignal \textbf{M}odulation): we write a single polysemous $v^*$ into the model and let the model's own downstream temporal modulation route the representation to time-correct answers---requiring no architectural changes, no auxiliary modules, and no per-period storage.

Our main contributions are as follows:
\begin{itemize}
\item We find through causal tracing on $(s,r,o,t)$ tuples that large language models process temporal knowledge via a two-stage account: time-agnostic subject retrieval at early MLP layers, followed by later-layer temporal modulation of the same subject signal (\S\ref{sec:trace}).
\item We formalize \emph{temporal polysemy} and propose \textbf{PRISM Edit}, which reconceptualizes temporal editing from writing separate per-period facts to writing a single polysemous $v^*$ that the model's own modulation disambiguates---requiring no architectural changes or auxiliary modules (\S\ref{sec:polysemy}).
\item We introduce \textsc{TimeConflict}, a temporal editing benchmark spanning 24 relations and 22{,}708 records, and show that PRISM Edit achieves overall state-of-the-art performance on core temporal metrics across \textsc{TimeConflict} and temporally augmented \textsc{CounterFact},  while being more than $2\times$ faster on temporal editing tasks (\S\ref{sec:experiments}).
\end{itemize}



\section{Related Work}
\subsection{Locate-then-Edit Knowledge Editing}
\label{sec:related_work}
Knowledge editing enables efficient updates of specific facts in LLMs without full retraining~\cite{zhang2024comprehensivestudyknowledgeediting}. Among various approaches, the locate-then-edit paradigm has proven particularly effective by first identifying where knowledge is stored and then directly modifying the corresponding parameters. ROME~\cite{meng2022locating} applies causal tracing to locate factual associations in mid-layer MLP modules and performs rank-one weight updates. MEMIT~\cite{meng2023memit} extends this to batch editing by distributing updates across multiple critical layers. AlphaEdit~\cite{fang2025alphaedit} further projects updates into the null space of preserved knowledge to minimize interference with unrelated facts. Other paradigms such as external memory approaches~\cite{hartvigsen2023grace,mitchell2022fast} and dual-memory architectures~\cite{wang2024wise} avoid direct parameter modification but introduce additional inference overhead. However, all these methods treat knowledge as temporally static, ignoring that historical facts remain valid within their original time scope.
\subsection{Temporal Knowledge Editing}
Real-world knowledge evolves over time, yet most editing methods treat facts as temporally static, simply overwriting old facts with new ones. Early work probes this temporal dimension on the language modeling side: \citet{dhingra-etal-2022-time} introduce TempLAMA and show that LMs encode temporally scoped facts but quickly become stale, motivating time-aware updates. \citet{yin2024history} first formalized temporal knowledge editing (TKE), arguing that historical facts remain valid within their original time scope and proposing the AToKe benchmark. Subsequent work has explored this direction from various angles: Temple-MQA~\cite{cheng2024multi} constructs temporal knowledge graphs for time-aware multi-hop reasoning, and SPIKE~\cite{zhao2026multi} introduces sparse parameter injection anchored to temporal markers for multi-granularity temporal editing. Our method, PRISM Edit, takes a fundamentally different approach: we jointly optimize a single time-agnostic $v^*$, allowing the model to naturally produce different answers based on temporal context.

\begin{figure*}[t]
\centering
\includegraphics[width=\textwidth]{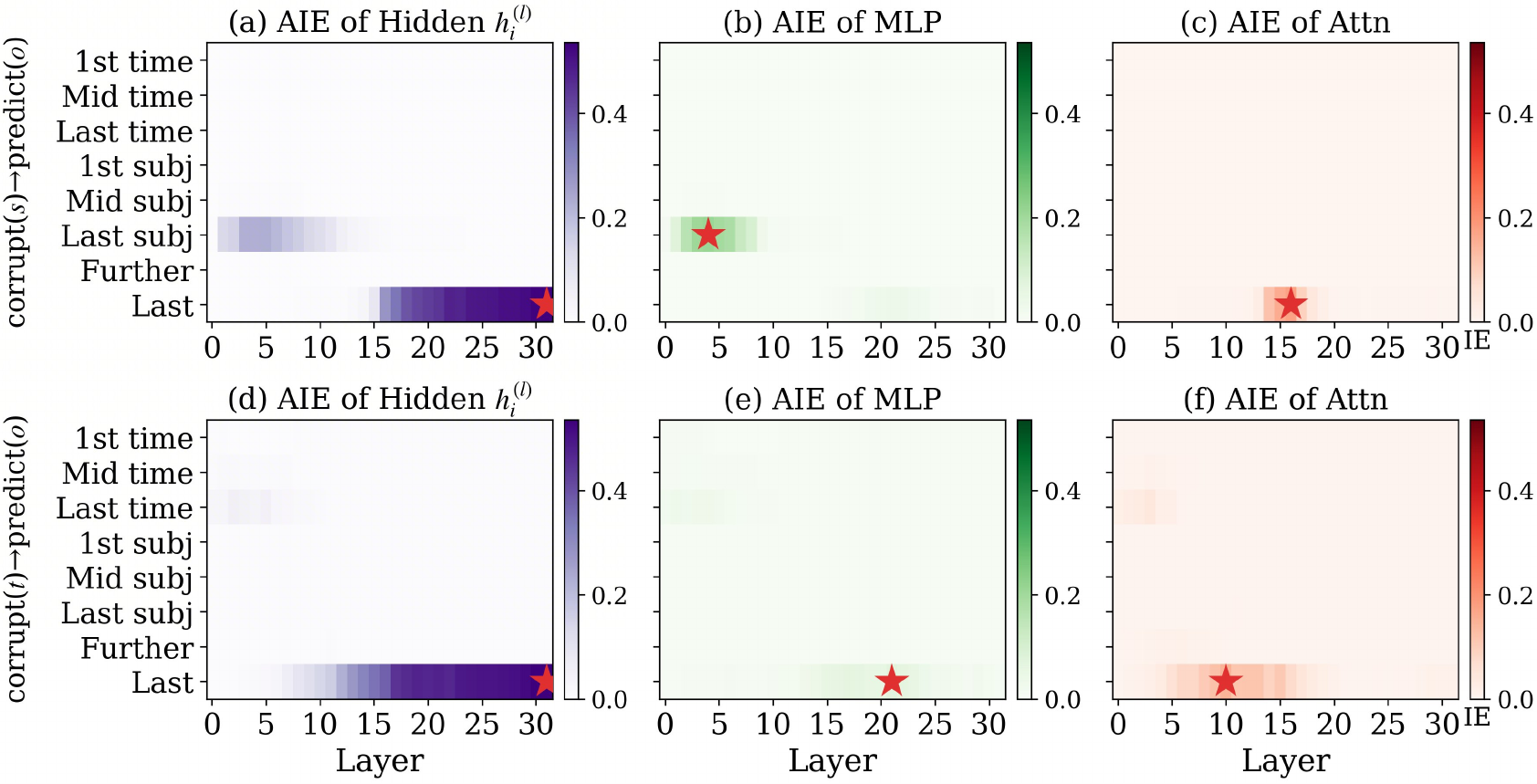}
\caption{\textbf{Average indirect effect (AIE) of individual model components over 500 temporal facts reveals two spatially separated causal sites.} Top row corrupts subject tokens; bottom row corrupts time tokens. \textbf{(a)}~An early site emerges at the subject-last token and a late site at the last token. \textbf{(b)}~MLP dominates the early site, reflecting abstract subject knowledge retrieval. \textbf{(c)}~Attention dominates the late site, reflecting subject-to-prediction transport. \textbf{(d)}~When time tokens are corrupted, the early site at the time-last token is weak, but a strong late site at the last token confirms that temporal context critically influences the final prediction. \textbf{(e)}~MLP effects at the subject position largely vanish; a mild MLP peak at the last token reflects temporal processing after the subject signal arrives. \textbf{(f)}~Attention carries temporal context to the last token earlier than subject information.}
\label{fig:dual_pathway}
\end{figure*}
\section{Tracing Temporal Knowledge in LLMs}
\label{sec:trace}


\begin{figure}[t]
\centering
\includegraphics[width=\columnwidth]{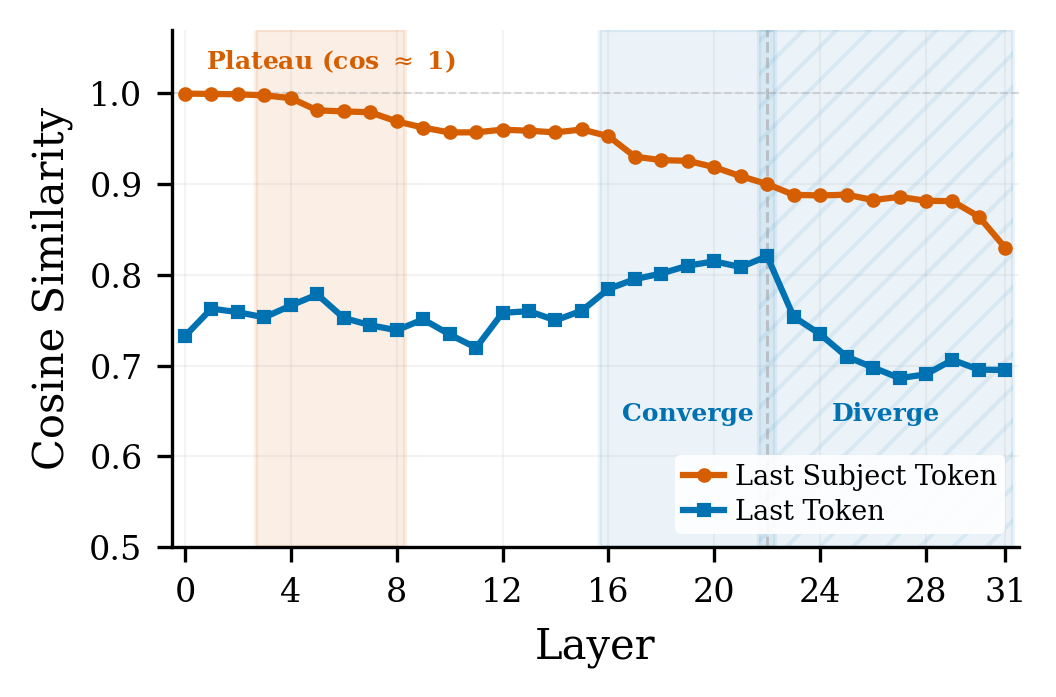}
\caption{Layer-wise cosine similarity of hidden states under new-time vs.\ old-time prompts.}
\label{fig:unedited_cos}
\end{figure}

In this section, we investigate how large language models internally process temporal knowledge through causal tracing and representational similarity analysis.

\subsection{Methods}
\label{sec:trace_method}

Understanding how temporal context influences factual retrieval requires tracing information flow across the model's internal layers. To this end, we adopt \textit{causal tracing} \cite{meng2022locating}, a probing technique that measures the causal contribution of individual hidden states to a model's prediction. The procedure involves three runs: (1)~a \textit{clean run} that records all hidden states under uncorrupted inputs, (2)~a \textit{corrupted run} that adds Gaussian noise to a selected token group at the embedding layer, yielding a degraded prediction, and (3)~a \textit{restore run} that restores a single (token, layer, component) activation to its clean-run value, while leaving all other activations at their corrupted values. 
From this, the indirect effect (IE) 
is defined as $P_{\text{restore}}(\text{correct answer}) - P_{\text{corrupted}}(\text{correct answer})$, 
measuring how much restoring a single activation recovers the model's prediction, thereby identifying which activations are causally responsible for factual recall.

For standard $(s, r, o)$ triplets, causal tracing has localized 
factual recall to the early MLP layers~\cite{geva-etal-2021-transformer} at the subject's last token, where the indirect effect peaks. Temporal knowledge, however, follows an extended $(s,r,o,t)$ structure, where producing the correct answer demands not only recognizing the subject but also anchoring in time. This raises a key question: 
\textbf{how do subject and temporal signals jointly determine the model's prediction?}

To answer this, we extend causal tracing to disentangle subject and temporal contributions. Rather than corrupting subject tokens as a whole, we corrupt subject tokens and time tokens separately, obtaining individual IE maps for each information pathway. Concretely, we compute the average IE on LLaMA-3-8B-Instruct \cite{grattafiori2024llama} over 500 temporal knowledge pairs that the model answers correctly, varying the choice of mediator over all (token, layer, component) triplets with component $\in$ \{MLP, Attention\}. See Appendix~\ref{app:ex_trace} for more details. 

Additionally, we perform a representational similarity analysis: for the same subject under new-time vs.\ old-time prompts, we compute layer-wise cosine similarity of hidden states at key token positions, revealing how temporal information progressively affects representations across layers.

\subsection{Findings}
\label{sec:trace_findings}

The resulting AIE maps (Figure~\ref{fig:dual_pathway}) show a striking asymmetry: subject effects concentrate at the early MLP layers (L4--8) of the subject-last token, while temporal effects are more distributed across positions and components. Combined with the representational similarity analysis (Figure~\ref{fig:unedited_cos}), these maps reveal three key findings on the model's temporal processing mechanism:

\paragraph{Finding 1: Subject retrieval is time-agnostic.}
\label{sec:time_agnostic}
Consistent with prior work on factual recall~\citep{meng2022locating, meng2023memit, dai-etal-2022-knowledge, geva-etal-2023-dissecting}, our causal tracing confirms the early MLP layers (L4--8) at the last subject token as the core knowledge-storage site, with the largest indirect effect peaking at L4 MLP (IE=0.549; Figure~\ref{fig:dual_pathway}b). What is new, however, is that this abstract knowledge retrieval is largely time-agnostic, supported by two converging pieces of evidence: (i)~the last subject representation forms a plateau ($\cos \geq 0.97$) across the critical MLP layers (L4--8) (Figure~\ref{fig:unedited_cos}, orange), showing that the subject representation is 
nearly invariant to different temporal context; (ii)~when time tokens are corrupted (Figure~\ref{fig:dual_pathway}e), the IE at the last subject early MLP layers remains negligible, indicating that temporal information rarely enters the subject pathway. 

\paragraph{Finding 2: Temporal context arrives early but acts late.}
\label{sec:time_modulates}
To the last token, temporal information arrives earlier than subject information: a dedicated attention route, starting near L5 and peaking around L10--15, carries temporal context to the prediction position (IE=0.356 at L10, Figure~\ref{fig:dual_pathway}f), several layers before the subject signal is transported to the last token via L16 attention (Figure~\ref{fig:dual_pathway}c). Yet its \emph{effect} on the output does not manifest immediately. Figure~\ref{fig:unedited_cos} (blue) shows that the two temporal prompts first \emph{converge} in the middle layers (cosine rising to $\approx 0.82$). This convergence is driven by the subject signal: arriving at the last token via L16 attention, the (shared) subject representation momentarily dominates the hidden state, 
masking the temporal differences between the two prompts. This attention-mediated temporal pathway is also supported by the recently identified Temporal Heads \citep{park-etal-2025-time}. Only in the upper layers do the two prompts \emph{diverge} (cosine dropping to $\approx 0.70$), reflecting temporal modulation that acts only after the subject signal is in place. 

Together with Finding~1, this is consistent with a \emph{two-stage} account: the subject signal arrives and dominates the last-token representation (convergence); the temporal signal, although it arrived earlier, modulates this subject-grounded signal in the later layers (divergence).

We further probe this route with attention-route severing on the PRISM-edited model (Appendix~\ref{app:severing}) to pinpoint where temporal context is causally necessary. Although tracing reveals active temporal transport through L9--15, only the earliest segment proves necessary for modulating the subject representation: severing L2--8 collapses TC from \(74.0\) to \(23.0\) while leaving CRS unchanged, whereas severing the peak segment (L9--15) or later layers (L16--31) leaves TC intact. Together, tracing and severing sharpen the two-stage account by showing that the causal necessity of temporal context lies earlier than its transport peak.

\paragraph{Finding 3: The editing site cannot discriminate time.}
\label{sec:key_insight}
Findings~1 and~2 together expose a fundamental obstacle to locate-and-edit methods (see Section~\ref{sec:related_work}) on temporal facts. Since these methods edit the MLP at the subject-last token, where $k$ is the input that conditions the write, any linear update $\Delta W$ at this layer satisfies
\begin{equation}
\|\Delta W\, k_{\text{new}} - \Delta W\, k_{\text{old}}\|
\;\leq\; \|\Delta W\| \cdot \underbrace{\|k_{\text{new}} - k_{\text{old}}\|}_{\approx\, 0}.
\label{eq:edit-collapse}
\end{equation}
where $k_{\text{new}}$ and $k_{\text{old}}$ denote the MLP input at the subject-last token under the new-time and old-time prompts, respectively. In other words, the two temporal prompts cannot be used as distinct retrieval conditions, so a single edit cannot deliver different values $v_{\text{new}}$ and $v_{\text{old}}$ for them. This is a geometric limit of the representation space, not a failure of optimization---no choice of 
optimizer or objective can recover what the keys cannot distinguish. And it is exactly why methods that try to edit different answers for different times, such as METO~\cite{yin2024history}, fail at this layer.

\paragraph{Design implication for editing.}
Taken together, the three key findings describe a coherent intrinsic mechanism: \emph{one time-agnostic subject signal plus later temporal modulation yields different answers}. This naturally suggests the right design principle: instead of forcing the edit layer to separate time---something it cannot do, as Equation~\ref{eq:edit-collapse} shows---we should let the model's own temporal modulation pathway handle the disambiguation. The next section formalizes this principle and turns it into a concrete editing objective.

\section{PRISM Editing: Aligning Edits with Temporal Modulation}
\label{sec:polysemy}

The mechanistic trace in \S\ref{sec:trace} reveals a structural separation between \emph{where} a factual edit is stored and \emph{how} the model resolves its temporal meaning. We now formalize this separation and show why it causes existing methods to fail, then derive PRISM Edit as a direct remedy.

\subsection{Background: Locate-then-Edit Formulation}
\label{sec:formalization}

Locate-then-edit methods~\citep{meng2022locating, meng2023memit, fang2025alphaedit} first identify an editing site, then compute a target hidden state at that site and write it into the weights. Let $\boldsymbol{v}_0$ denote the unedited hidden state at hook layer $\ell^\star$ and subject-last position, and $G_{\boldsymbol{v}_0+\delta}$ denote the model with that position replaced by $\boldsymbol{v}_0+\delta$. These methods optimize $\delta$ on a single edit prompt, in which any time expression is treated as ordinary 
context:
\begin{equation}
\label{eq:standard_v}
  \begin{aligned}
  \boldsymbol{\delta}^{*}_{\mathrm{std}}
  &=
  \arg\min_{\delta}
  \Bigl[
    -\log \mathbb{P}_{G_{\boldsymbol{v}_0+\delta}}(o_\text{edit}\mid p_\text{edit})
    + \Omega(\delta)
  \Bigr],\\
  \boldsymbol{v}^{*}_{\mathrm{std}}
  &= \boldsymbol{v}_0 + \boldsymbol{\delta}^{*}_{\mathrm{std}} .
  \end{aligned}
\end{equation}
Here $p_\text{edit}$ is the edit prompt built from $(s,r)$ and, when applicable, a temporal context $t_\text{edit}$; $o_\text{edit}$ is the target answer, and $\Omega(\delta)$ is a locality regularizer. The optimized $\boldsymbol{v}^{*}_{\mathrm{std}}$ is then realized as weight updates $\{\Delta W_\ell\}_{\ell \in \mathcal{L}}$ at a chosen set of subject-last MLP layers $\mathcal{L}$.
By Finding~1, the resulting updates are time-agnostic at the editing site---collapsing every temporal query to the 
same target $\boldsymbol{v}^{*}_{\mathrm{std}}$ 
(Eq.~\ref{eq:edit-collapse}). To break this collapse, PRISM Edit (\S\ref{sec:method}) reformulates the target computation step.

\subsection{PRISM Edit}
\label{sec:method}
\paragraph{Downstream temporal modulation.}
Although the edit overwrites the subject-position state with 
the time-blind target $\boldsymbol{v}^{*}_{\mathrm{std}}$, 
temporal information is not lost: by Finding~2, higher-layer 
attention at the last token combines the propagated subject 
representation with time-token hidden states before 
producing the final prediction. Schematically,
\begin{equation}
\label{eq:routing}
\mathbf{h}_{\mathrm{out}}^{\mathrm{last}}(s, t) 
\;=\; f\bigl(\bar{\mathbf{s}},\, \mathbf{c}_t\bigr),
\end{equation}
where $\mathbf{h}_{\mathrm{out}}^{\mathrm{last}}(s, t)$ is the final last-token residual state, $\bar{\mathbf{s}}$ is the subject-position representation propagated from the 
editing site (nearly shared across temporal contexts by Finding~1), $\mathbf{c}_t$ are the hidden states at time-token positions, and $f$ aggregates contributions from layers above the editing range. Different $\mathbf{c}_t$ can elicit different outputs from a shared $\bar{\mathbf{s}}$
---a latent routing capacity that PRISM Edit will exploit.
\paragraph{Temporal polysemy.}
This routing capacity motivates what we call \emph{temporal polysemy}: one written value should support multiple context-conditioned readouts. Rather than storing independent answer representations for each timestamp, a single $\boldsymbol{v}^{*}$ should serve as a shared subject anchor that downstream temporal modulation reads differently under different time cues:
\[
  \boldsymbol{v}^{*}
  \;\xrightarrow{\;\mathbf{c}_{t_1}\;}\; o_1,
  \qquad
  \boldsymbol{v}^{*}
  \;\xrightarrow{\;\mathbf{c}_{t_2}\;}\; o_2,
  \qquad \ldots
\]
This mirrors linguistic polysemy---one form, multiple context-dependent meanings. The required mechanism already exists in the base model (Section~\ref{sec:trace}). PRISM therefore adds no explicit router; instead, it optimizes the written value so that the existing attention pathway can read $\boldsymbol{v}^{*}$ under the available temporal cue $\mathbf{c}_t$.

\paragraph{Joint optimization objective.}
To realize temporal polysemy, $\boldsymbol{v}^{*}=
\boldsymbol{v}_0+\boldsymbol{\delta}^{*}$ must be 
compatible with \emph{every} temporal context 
simultaneously. Whereas prior work mainly refines the weight-write stage, PRISM Edit revisits the target-computation stage itself: rather than optimizing $\boldsymbol{\delta}$ from a single edit prompt as in Eq.~\eqref{eq:standard_v}, PRISM 
learns one shared $\boldsymbol{\delta}^{*}$ from all 
relevant temporal contexts. Given $N$ temporal contexts $\{(t_i, o_i)\}_{i=1}^{N}$ for the same subject--relation pair $(s,r)$, we solve:
\begin{equation}
\label{eq:general_loss}
  \begin{aligned}
  \mathcal{L}_i(\delta)
  &=
  -\log \mathbb{P}_{G_{\boldsymbol{v}_0+\delta}}
  [\,o_i \mid p_i(s,r,t_i)\,],\\
  \boldsymbol{\delta}^{*}
  &= \arg\min_{\delta}
  \sum_{i=1}^{N}\lambda_i\,\mathcal{L}_i(\delta)+\Omega(\delta),\\
  \boldsymbol{v}^{*}
  &=\boldsymbol{v}_0+\boldsymbol{\delta}^{*}.
  \end{aligned}
\end{equation}
where $p_i(s,r,t_i)$ is a prompt with subject $s$, relation $r$, and temporal context $t_i$; the log-probability is summed over all target tokens of $o_i$; $\lambda_i$ are per-context weights, and $\Omega(\delta)$ collects locality and norm regularizers. This formulation naturally extends to an arbitrary number of temporal snapshots.

The most common temporal editing case is a two-state 
transition with $(t_\text{new}, t_\text{old})$, for which 
Eq.~\eqref{eq:general_loss} instantiates as:
\begin{equation}
\label{eq:joint_loss}
  \begin{aligned}
  \boldsymbol{\delta}^{*} = \arg\min_{\delta}\;
  &\mathcal{L}_\text{new}
  + \lambda_\text{old}\mathcal{L}_\text{old}
  + \mathcal{L}_\text{bare}\\
  &+ \Omega(\delta),\\
  \boldsymbol{v}^{*}
  &=\boldsymbol{v}_0+\boldsymbol{\delta}^{*}.
  \end{aligned}
\end{equation}
$\mathcal{L}_\text{new}$ and $\mathcal{L}_\text{old}$ are the NLL losses for $(t_\text{new}, o_\text{new})$ and 
$(t_\text{old}, o_\text{old})$, and $\mathcal{L}_\text{bare}$ targets $o_\text{new}$ on a prompt without temporal context---reflecting that real-world queries often omit explicit time markers, in which case the model should default to the current answer.

The complete PRISM Edit procedure is summarized in Algorithm~\ref{alg:prism}.



\section{Experiments}
\label{sec:experiments}



\begin{table*}[t]
\centering
\small
\begin{tabular}{ll|cccccc}
\toprule
\textbf{Dataset} & \textbf{Method} & \textbf{CES}$\uparrow$ & \textbf{CES-P}$\uparrow$ & \textbf{CRS}$\uparrow$ & \textbf{HES}$\uparrow$ & \textbf{HES-P}$\uparrow$ & \textbf{TC}$\uparrow$ \\
\midrule
& Pre-Edit & \meanstd{1.1}{0.3} & \meanstd{0.6}{0.1} & \meanstd{1.0}{0.3} & \meanstd{48.7}{1.6} & \meanstd{46.6}{1.2} & \meanstd{0.4}{0.2} \\
\cmidrule(l){2-8}
& Fine-Tuning & \meanstd{49.5}{1.6} & \meanstd{38.1}{1.3} & \meanstd{13.7}{1.1} & \meanstd{62.1}{1.5} & \meanstd{54.8}{1.4} & \meanstd{33.1}{1.5} \\
& ROME & \meanstd{0.0}{0.0} & \meanstd{0.0}{0.0} & \meanstd{0.0}{0.0} & \meanstd{0.1}{0.0} & \meanstd{0.1}{0.0} & \meanstd{0.0}{0.0} \\
\textsc{CounterFact}$^{\dagger}$ & MEMIT & \meanstd{47.3}{1.6} & \meanstd{38.4}{1.3} & \meanstd{35.5}{1.5} & \meanstd{60.9}{1.5} & \meanstd{56.7}{1.3} & \meanstd{20.2}{1.3} \\
& AlphaEdit & \meanstdu{64.5}{1.5} & \meanstdu{51.6}{1.2} & \meanstd{33.6}{1.5} & \meanstdu{80.6}{1.3} & \meanstdu{74.4}{1.1} & \meanstdu{46.8}{1.6} \\
& METO & \meanstd{59.4}{1.6} & \meanstd{51.3}{1.6} & \meanstdu{36.7}{1.5} & \meanstd{72.3}{1.4} & \meanstd{68.2}{1.5} & \meanstd{35.0}{1.5} \\
& \textbf{PRISM Edit} & \meanstdb{86.4}{1.1} & \meanstdb{75.5}{1.0} & \meanstdb{82.5}{1.2} & \meanstdb{95.5}{0.7} & \meanstdb{85.1}{0.7} & \meanstdb{82.4}{1.2} \\
\midrule
& Pre-Edit & \meanstd{9.5}{0.9} & \meanstd{7.5}{0.8} & \meanstd{9.2}{0.9} & \meanstd{15.7}{1.2} & \meanstd{13.9}{1.1} & \meanstd{2.9}{0.5} \\
\cmidrule(l){2-8}
& Fine-Tuning & \meanstd{12.8}{1.1} & \meanstd{10.9}{1.0} & \meanstd{10.7}{1.0} & \meanstd{11.4}{1.0} & \meanstd{9.0}{0.9} & \meanstd{4.0}{0.6} \\
& ROME & \meanstd{0.0}{0.0} & \meanstd{0.0}{0.0} & \meanstd{0.0}{0.0} & \meanstd{0.0}{0.0} & \meanstd{0.0}{0.0} & \meanstd{0.0}{0.0} \\
\textsc{TimeCF-Factual} & MEMIT & \meanstd{11.9}{1.0} & \meanstd{9.2}{0.9} & \meanstd{14.9}{1.1} & \meanstd{13.4}{1.1} & \meanstd{8.2}{0.9} & \meanstd{4.0}{0.6} \\
& AlphaEdit & \meanstdu{43.6}{1.6} & \meanstdu{32.2}{1.5} & \meanstdu{34.1}{1.5} & \meanstdu{49.6}{1.6} & \meanstdu{36.5}{1.5} & \meanstdu{19.4}{1.3} \\
& METO & \meanstd{30.5}{1.5} & \meanstd{21.9}{1.3} & \meanstd{20.4}{1.3} & \meanstd{39.7}{1.5} & \meanstd{28.3}{1.4} & \meanstd{13.1}{1.1} \\
& \textbf{PRISM Edit} & \meanstdb{59.2}{1.6} & \meanstdb{48.1}{1.6} & \meanstdb{61.7}{1.5} & \meanstdb{60.8}{1.5} & \meanstdb{51.1}{1.6} & \meanstdb{37.4}{1.5} \\
\midrule
& Pre-Edit & \meanstd{1.6}{0.4} & \meanstd{1.2}{0.3} & \meanstd{2.0}{0.4} & \meanstd{15.7}{1.2} & \meanstd{13.9}{1.1} & \meanstd{0.5}{0.2} \\
\cmidrule(l){2-8}
& Fine-Tuning & \meanstd{9.0}{0.9} & \meanstd{7.1}{0.8} & \meanstd{9.6}{0.9} & \meanstd{13.2}{1.1} & \meanstd{10.9}{1.0} & \meanstd{2.4}{0.5} \\
& ROME & \meanstd{0.0}{0.0} & \meanstd{0.0}{0.0} & \meanstd{0.0}{0.0} & \meanstd{0.0}{0.0} & \meanstd{0.1}{0.1} & \meanstd{0.0}{0.0} \\
\textsc{TimeCF-Fictional} & MEMIT & \meanstd{6.0}{0.8} & \meanstd{4.9}{0.7} & \meanstd{6.1}{0.8} & \meanstd{9.8}{0.9} & \meanstd{8.3}{0.9} & \meanstd{1.7}{0.4} \\
& AlphaEdit & \meanstdu{16.0}{1.2} & \meanstdu{9.7}{0.9} & \meanstdu{9.6}{0.9} & \meanstdu{48.8}{1.6} & \meanstdu{38.4}{1.5} & \meanstdu{6.5}{0.8} \\
& METO & \meanstd{9.2}{0.9} & \meanstd{4.6}{0.7} & \meanstd{4.2}{0.6} & \meanstd{35.9}{1.5} & \meanstd{27.0}{1.4} & \meanstd{4.6}{0.7} \\
& \textbf{PRISM Edit} & \meanstdb{36.0}{1.5} & \meanstdb{25.4}{1.4} & \meanstdb{37.4}{1.5} & \meanstdb{60.2}{1.5} & \meanstdb{49.0}{1.6} & \meanstdb{22.8}{1.3} \\
\bottomrule
\end{tabular}
\caption{Main results on temporal knowledge editing 
(1{,}000 records per dataset, sequential editing in 
batches of 100). Best in \textbf{bold}, second-best 
\underline{underlined}. Metric definitions are in 
\S\ref{sec:exp_setup}; $^{\dagger}$ denotes our temporal augmentation of 
\textsc{CounterFact}~\citep{meng2022locating} 
(Appendix~\ref{app:TimeConflict}).}
\label{tab:main}
\end{table*}

\subsection{Experimental Setup}

\paragraph{Task Definition.}
We study \emph{temporally conflicting edits}: given a subject--relation pair whose object changes over time, the edited model must (i) return the new object under a post-update timestamp, (ii) return the old object under a pre-update timestamp, and (iii) default to the new object under a \emph{bare} prompt without temporal context.

\paragraph{Datasets.}
Static editing benchmarks such as \textsc{CounterFact}~\cite{meng2022locating} 
and zsRE~\cite{levy2017zero} cast an edit as a single $(s,r,o\!\rightarrow\!o^*)$ rewrite, with no temporal dimension. The closest temporal alternative, ATOKE~\citep{yin2024history}, 
covers only 9 YAGO3 relations at year-level \emph{since/until} granularity---too coarse for mid-year role changes and too narrow for relational generality. 

We therefore introduce \textbf{\textsc{TimeConflict}} (\textsc{TimeCF}), a temporal editing dataset of 22{,}708 records over 24 relations with \emph{day-level} time stamps---broader in coverage and finer in granularity. Each record has the form $\langle s, r, o_{\text{before}}, t_{\text{before}}, o_{\text{after}}, t_{\text{after}} \rangle$ and comes in two variants: \textsc{TimeCF-Factual} keeps the real successor for evaluation under real-world updates, while \textsc{TimeCF-Fictional} replaces 
$o_{\text{after}}$ with a counterfactual entity, eliminating contamination from pretraining knowledge. To assess generalization, we additionally evaluate on \textsc{CounterFact}$^{\dagger}$, a temporally augmented version of \textsc{CounterFact}~\citep{meng2022locating}. Dataset details are in Appendix~\ref{app:TimeConflict}.

\paragraph{Metrics.}
\label{sec:exp_setup}
We adopt and extend the temporal-edit metrics of \citet{yin2024history}. All metrics are computed on the post-edit model. 
\begin{itemize}
\item \textbf{CES} / \textbf{HES} (\textbf{C}urrent / 
\textbf{H}istorical \textbf{E}xplicit-time \textbf{S}core): 
accuracy on post-/pre-update timestamp prompts;
\item \textbf{CES-P} / \textbf{HES-P}: paraphrased variants 
of CES / HES;
\item \textbf{CRS} (\textbf{C}urrent \textbf{R}elative-time 
\textbf{S}core): accuracy on \emph{bare} prompts without 
explicit time markers, where the model should default to 
the new object;
\item \textbf{TC} (\textbf{T}emporal \textbf{C}onsistency): 
the fraction of records on which CES and HES are jointly 
correct---the strictest measure of current--historical 
coexistence in a single model.
\end{itemize}
Unlike prior work that evaluates edits via token-level negative log-likelihood (NLL), we adopt normalized substring match as our primary metric—a strictly harder criterion that requires the model to generate the correct entity rather than merely assigning it higher probability among candidates.


\paragraph{Models and Baselines.}

We conduct experiments on four backbones: \textbf{Llama-3}(8B)~\citep{grattafiori2024llama}, \textbf{GPT-J}(6B)~\citep{wang2021gptj}, \textbf{GPT-2 XL} (1.5B)~\citep{radford2019language}, and \textbf{Qwen3-4B-Instruct}~\citep{qwen3}. Results in the main text use Llama-3;results on the other three backbones are reported in Appendix~\ref{app:other_ex}.  

We compare two families of editors: \textbf{(i) locate-then-edit}---ROME~\cite{meng2022locating}, MEMIT~\cite{meng2023memit}, AlphaEdit~\cite{fang2025alphaedit} and METO~\cite{yin2024history}, where the first three do not account for temporal information and METO additionally incorporates temporal context into the editing objective; \textbf{(ii) memory-based}---GRACE~\cite{hartvigsen2023grace} and WISE~\cite{wang2024wise}. Memory-based methods introduce \emph{auxiliary memory modules} 
with retrieval/routing mechanisms rather than directly editing the original weights at a target site, making them not directly comparable to (i); we report their results in Appendix~\ref{app:other_ex}. We further include \textbf{Fine-Tuning} (FT) and the unedited model (\textbf{Pre-Edit}) as references. For methods that do not explicitly model time, we treat the new and historical facts as two independent edit requests. 

\paragraph{Implementation Details.}
Following standard locate-then-edit practice~\citep{meng2022locating,meng2023memit,fang2025alphaedit}, we write edits into early-to-mid MLP layers: L4--8 for LLaMA-3 and Qwen3-4B, L3--8 for GPT-J, and L13--17 for GPT-2 XL; for LLaMA-3 and GPT-J these windows coincide with the subject-MLP peaks identified by causal tracing (\S\ref{sec:trace}). The target vector $\boldsymbol{v}^{*}$ is optimized for 25 gradient steps (20 for GPT-2 XL) with learning rate $0.1$ ($0.5$ for GPT-J and GPT-2 XL), weight decay $0.5$, and KL factor $0.0625$; the loss weights are $\lambda_\text{old}{=}\lambda_\text{bare}{=}2$, the best trade-off in our ablation (Table~\ref{tab:ablation_loss}). All supplementary analyses reuse this configuration without additional tuning.

\subsection{Main Results}

Table~\ref{tab:main} reports LLaMA-3 results on three datasets. PRISM Edit achieves the best scores on all six metrics, with the largest gains on \textbf{CRS} and \textbf{TC}---the indicators most directly testing temporal coexistence. We highlight three observations.

\paragraph{Uniform gains across temporal conditions.}
Averaged across the three datasets, PRISM Edit outperforms the strongest baseline by $+19.2$ on CES, $+33.7$ on CRS, and $+12.5$ on HES---both current and historical facts are recalled accurately, with the largest gain on bare prompts.  We attribute this CRS lead to the fact that, without an explicit temporal anchor, the model tends to fall back to the default behavior at the edited site; PRISM Edit makes this default an explicit optimization target, whereas conventional editing methods can only resolve it implicitly.

\paragraph{Joint correctness under temporal coexistence.}
\textbf{TC} requires CES and HES to be jointly correct on the same record. PRISM Edit improves over the second-best method by $+23.3$ on TC on average across the three datasets, indicating not isolated per-query wins but genuine coexistence of current and historical answers within one edited model.

\paragraph{Faster editing without a quality trade-off.}
Table~\ref{tab:edit_time} shows PRISM Edit averages $5.44$\,s per record---$2.8\times$ faster than AlphaEdit and $5.2\times$ faster than METO, while achieving the best scores in Table~\ref{tab:main}. The speedup comes from both the editing protocol and the optimization: PRISM Edit treats one record as a single edit rather than two independent edits per temporal condition (four for METO), and the multi-time target provides stronger supervision when solving the shared value.

\begin{table}[t]
\centering
\small
\setlength{\tabcolsep}{3pt}
\resizebox{\columnwidth}{!}{%
\begin{tabular}{lcccccc}
\toprule
Method & FT$^{\ddagger}$ & ROME & METO & AlphaEdit & MEMIT & \textbf{PRISM Edit} \\
\midrule
Time (s) $\downarrow$ & 0.91 & 38.18 & 28.32 & 15.10 & 11.74 & \textbf{5.44} \\
\bottomrule
\end{tabular}}
\caption{Average per-record edit time on 
\textsc{LLaMA-3} (8B), averaged over 
\textsc{CounterFact}$^{\dagger}$, \textsc{TimeCF-Factual}, and \textsc{TimeCF-Fictional}. 
$^{\ddagger}$FT is faster but fails editing (Table~\ref{tab:main}).}
\label{tab:edit_time}
\end{table}



\subsection{Ablation Study}
\label{sec:analysis}

Compared with conventional editing objectives that optimize the new fact alone, PRISM Edit additionally introduces historical and bare-time targets. We ablate their weights on \textsc{TimeCF-Factual} (Table~\ref{tab:ablation_loss}, 
$\lambda_\text{new}{=}1$ fixed), to test whether 
$\mathcal{L}_\text{old}$ enables coexistence and 
$\mathcal{L}_\text{bare}$ controls the bare-prompt default.

\begin{table}[t]
\centering
\small
\resizebox{\columnwidth}{!}{%
\begin{tabular}{cc|cccccc|c}
\toprule
$\lambda_\text{old}$ & $\lambda_\text{bare}$ & CES & CES-P & HES & HES-P & CRS & TC & Avg \\
\midrule
\multicolumn{9}{l}{\textit{Without bare-prompt supervision ($\lambda_\text{bare}{=}0$):}} \\
1 & 0 & 76.0 & 53.0 & 67.0 & 49.0 & 45.0 & 50.0 & 56.7 \\
2 & 0 & 73.0 & 50.0 & 77.0 & 59.0 & 39.0 & \textbf{61.0} & 59.8 \\
3 & 0 & 64.0 & 49.0 & \textbf{78.0} & \textbf{62.0} & 36.0 & 56.0 & 57.5 \\
\midrule
\multicolumn{9}{l}{\textit{With bare-prompt supervision:}} \\
1 & 1 & \textbf{79.0} & 53.0 & 60.0 & 48.0 & \textbf{78.0} & 47.0 & 60.8 \\
1 & 2 & \textbf{79.0} & \textbf{55.0} & 57.0 & 48.0 & 72.0 & 46.0 & 59.5 \\
2 & 1 & \textbf{79.0} & \textbf{55.0} & 69.0 & 54.0 & 70.0 & 57.0 & 64.0 \\
\rowcolor{gray!15}
\textbf{2} & \textbf{2} & \textbf{79.0} & \textbf{55.0} & 72.0 & 54.0 & 74.0 & 59.0 & \textbf{65.5} \\
3 & 1 & 71.0 & 49.0 & 72.0 & 56.0 & 69.0 & 56.0 & 62.2 \\
3 & 2 & 73.0 & 50.0 & 74.0 & 54.0 & 76.0 & 55.0 & 63.7 \\
\bottomrule
\end{tabular}%
}
\caption{Loss-weight ablation on \textsc{TimeCF-Factual} (100 records, \%), with $\lambda_\text{new}{=}1$ fixed. Best per column in \textbf{bold}. Without $\mathcal{L}_\text{bare}$, CRS collapses below 45\%; adding it recovers CRS to 70--78\%. The shaded row ($\lambda_\text{old}{=}\lambda_\text{bare}{=}2$) is our default, achieving the highest Avg and the best overall trade-off.}
\label{tab:ablation_loss}
\end{table}

Three patterns emerge from Table~\ref{tab:ablation_loss}. \textbf{(i)} $\mathcal{L}_\text{old}$ enables current--historical coexistence but over-corrects when too strong: with $\lambda_\text{bare}{=}0$, raising $\lambda_\text{old}$ from 1 to 2 lifts HES from 67.0 to 77.0, but $\lambda_\text{old}{=}3$ drops CES to 64.0. \textbf{(ii)} $\mathcal{L}_\text{bare}$ controls the bare-prompt default: without it, CRS stays at 36--45; adding it recovers CRS to at least 69. \textbf{(iii)} $(\lambda_\text{old},\lambda_\text{bare}){=}(2,2)$ achieves the best overall trade-off, which we adopt as the default.

Additional context-template ablations are in 
Appendix~\ref{app:ctx_ablation}.



\section{Conclusion}
\label{sec:conclusion}
We presented \textbf{PRISM Edit}, a mechanism-aligned method for temporal knowledge editing. Through the first causal-tracing analysis of temporal facts, we identify a two-stage account: early MLP layers retrieve a time-agnostic subject signal, and upper-layer attention modulates this signal with temporal context. This geometry exposes a fundamental obstacle for prior locate-then-edit methods: they force time-conditioned writes at a single MLP site, contradicting the model's two-stage solution where temporal disambiguation happens downstream. Rather than forcing this separation at the edit site, PRISM Edit jointly optimizes a single polysemous hidden state $\boldsymbol{v}^{*}$ that the model's own downstream modulation routes to time-appropriate answers---requiring no architectural changes or external memory.

We also introduce \textsc{TimeCF}, a temporal editing benchmark of 22{,}708 records over 24 relations with day-level time stamps. On \textsc{TimeCF} and a temporally augmented \textsc{CounterFact}, PRISM Edit achieves overall state-of-the-art performance on core temporal metrics, while running over $2\times$ faster than prior editors.

More broadly, our findings support a simple principle: \emph{aligning edits with the model's intrinsic computation} is more effective than imposing external structure---a mechanism-first approach that may generalize to other 
context-conditioned tasks in various application fields, e.g., robotics~\citep{zhang2026eiir} and manufacturing~\citep{xu2026assembly}.
\section{Limitations}
\label{sec:limitations}

While PRISM Edit demonstrates strong performance on temporal knowledge editing, we acknowledge several limitations. First, although \textsc{TimeCF} provides day-level annotations, our main experiments operate at the year level; editing at finer granularities (month or day) remains untested and may require denser supervision. Second, our evaluation focuses on the current-vs-historical 
setting (one current and one historical answer per 
subject); scaling the joint optimization to $k>2$ 
coexisting temporal targets is left to future work. Finally, the mechanism we exploit---upper-layer attention modulating a shared subject representation via contextual cues---is likely not unique to time. Extending $\boldsymbol{v}^{*}$ to other forms of context-conditioned polysemy is a natural direction enabled by our framework.

\section*{Acknowledgments}
This work was supported by the National Natural Science Foundation of China (92467204) and the Shenzhen Major Undertaking Plan (CJGJZD20240729141702003).

\bibliography{custom}

\appendix

\section{Causal Tracing Details}
\label{app:ex_trace}

\begin{figure*}[t]
\centering
\includegraphics[width=\linewidth]{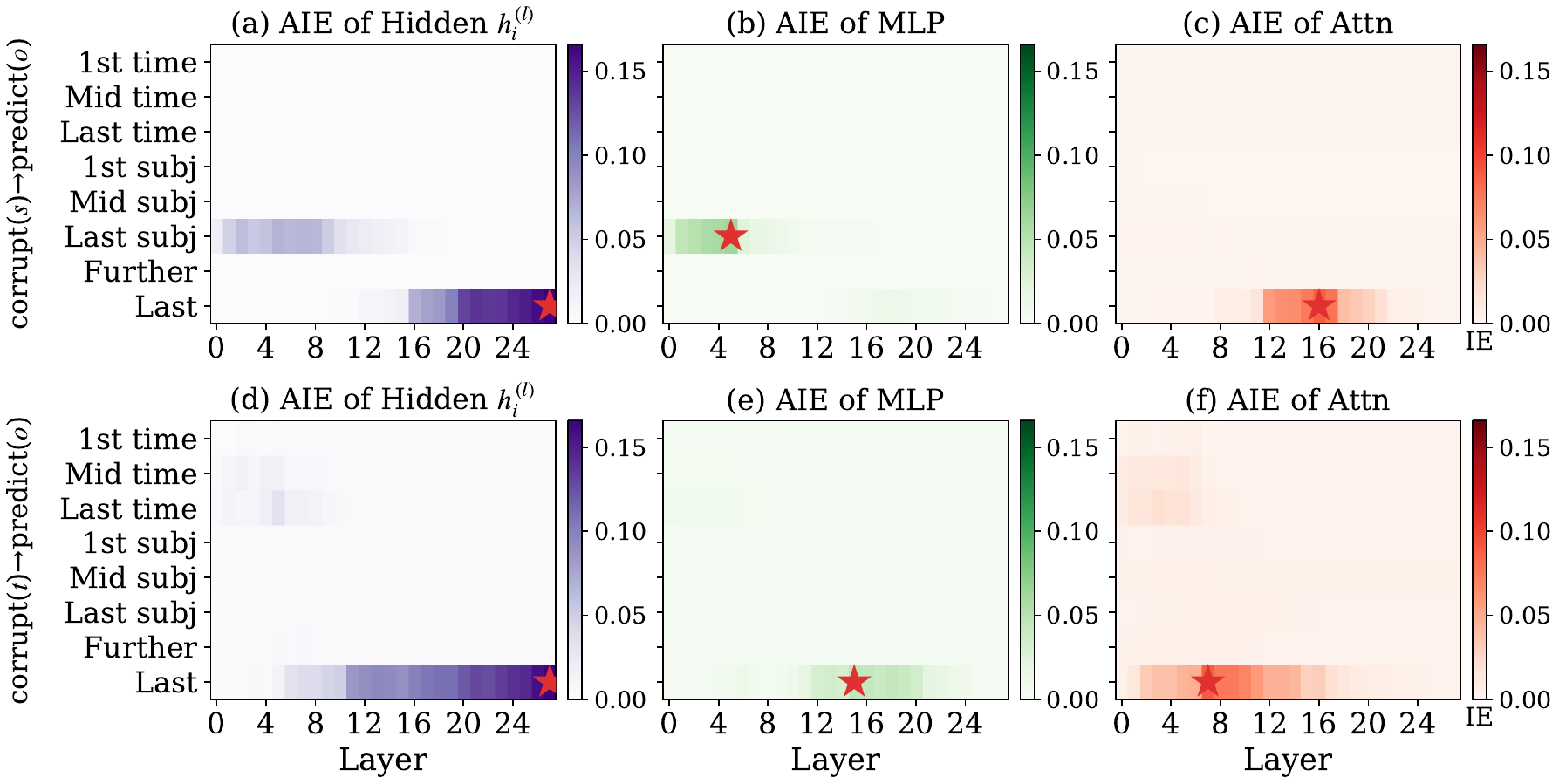}
\caption{Six-panel causal tracing heatmap for GPT-J, produced with the same protocol as Figure~\ref{fig:dual_pathway}.}
\label{fig:gptj_trace}
\end{figure*}

This section provides implementation details of the extended causal tracing procedure described in \S\ref{sec:trace}.  The base three-run protocol and indirect effect (IE) definition follow \citet{meng2022locating}; below we focus on our extensions.

\paragraph{Data filtering.}
We run the unedited model on all records in \textsc{TimeCF}-Factual and retain only those for which the model correctly predicts the target object under the time-conditioned prompt.  This filtering is necessary because causal tracing measures how much a restored activation \emph{recovers} the correct prediction; only when the model already produces the correct answer in the clean run can the indirect effect be meaningfully interpreted.  This yields 533 qualifying records.

\paragraph{Disentangling subject and temporal pathways.}
Our key extension is to \emph{separately} corrupt the subject tokens and the time tokens, yielding two independent IE maps that isolate each information pathway.

In the standard causal tracing protocol, the \emph{restore} step patches the full hidden state $\mathbf{h}_i^{(\ell)}$ at a given (token $i$, layer $\ell$) back to its clean-run value.  To further disentangle the contributions of different components, we perform \emph{component-level restoration}: in a Transformer layer the residual update decomposes as $\mathbf{h}_i^{(\ell)} = \mathbf{h}_i^{(\ell-1)} + \mathbf{a}_i^{(\ell)} + \mathbf{m}_i^{(\ell)}$, where $\mathbf{a}_i^{(\ell)}$ is the Attention sub-layer output and $\mathbf{m}_i^{(\ell)}$ is the MLP sub-layer output.  ``Restore MLP only'' means we patch $\mathbf{m}_i^{(\ell)}$ to its clean-run value while keeping $\mathbf{a}_i^{(\ell)}$ corrupted; ``Restore Attention only'' does the converse.  This allows us to measure each component's independent causal contribution to the final prediction.

Crossing two corruption targets (subject tokens vs.\ time tokens) with three restoration granularities (total hidden state, MLP only, Attention only) yields the six conditions shown in Figure~\ref{fig:dual_pathway}.  Panels~(a)--(c) corrupt the subject tokens: panel~(a) restores the full hidden state to reveal the overall subject-IE map; panels~(b) and~(c) restore only MLP or only Attention, respectively, to isolate each component's individual contribution.  Panels~(d)--(f) mirror this design for the time tokens.
For each panel, we compute the IE at every (layer, token-position) cell and average across all qualifying samples, yielding the heatmaps presented in Figure~\ref{fig:dual_pathway}.

\paragraph{Cross-architecture generalization.}
We apply the same procedure to GPT-J(6B).  As shown in Figure~\ref{fig:gptj_trace}, the temporal pathway consistently activates at earlier layers than the subject pathway, corroborating the findings on LLaMA-3 reported in \S\ref{sec:trace}.

\section{Dataset Details}
\label{app:TimeConflict}

\subsection{TimeConflict}
\subsubsection{Dataset Construction}
We construct \textsc{TimeConflict} from the Wikidata JSON dump (\texttt{latest-all.json.bz2}).
We extract temporal triples for 100 candidate properties using qualifier timestamps (P580 start-time, P582 end-time, P585 point-in-time), group them by subject, and detect temporal conflicts---pairs where \texttt{after\_object} starts after the cutoff (January 1, 2023) and \texttt{before\_object} is its most recent predecessor with a different identity.
After removing records with missing labels and deduplicating by (\texttt{before\_object}, \texttt{after\_object}, \texttt{after\_start}), 24 properties yield $\geq$10 valid conflicts each, producing 22{,}708 records in total.
For \textsc{TimeCF}-Fictional, we replace each \texttt{after\_object} with a random entity from the same-relation entity pool (seed\,=\,42), preserving original timestamps.

\subsubsection{Dataset Statistics}
\label{app:statistics}

Table~\ref{tab:statistics} summarizes the key statistics of \textsc{TimeConflict}. The per-relation record counts and sampling distribution are detailed in Table~\ref{tab:sampled_dist}. The distribution is long-tailed---the top 3 relations account for approximately 72\% of all records, reflecting real-world knowledge update frequency.

\begin{table}
\centering
\small
\caption{Summary statistics of \textsc{TimeConflict}.}
\label{tab:statistics}
\begin{tabular}{@{}lr@{}}
\toprule
\textbf{Statistic} & \textbf{Value} \\
\midrule
Total records & 22,708 \\
Relations & 24 \\
Unique subjects & $>$18,000 \\
Granularity & Day-level \\
Cutoff date & 2023-01-01 \\
Before-object range & 1751-09 -- 2024-12 \\
After-object range & 2023-01 -- 2025-12 \\
Experiment sample & 1,000 \\
\bottomrule
\end{tabular}
\end{table}

\begin{table}
\centering
\caption{Per-relation distribution of the 1,000 sampled records.}
\label{tab:sampled_dist}
\footnotesize
\setlength{\tabcolsep}{3pt}
\begin{tabular}{@{}llrr@{}}
\toprule
\textbf{ID} & \textbf{Relation} & \textbf{\#Samp.} & \textbf{\#Full} \\
\midrule
P108  & employer & 231 & 8,747 \\
P39   & position held & 145 & 5,538 \\
P361  & part of & 74 & 75 \\
P276  & location & 70 & 70 \\
P35   & head of state & 66 & 66 \\
P54   & member of sports team & 57 & 2,167 \\
P710  & participant & 49 & 49 \\
P102  & political party & 44 & 44 \\
P166  & award received & 40 & 1,519 \\
P449  & original broadcaster & 35 & 35 \\
P131  & admin.\ territorial entity & 32 & 1,236 \\
P286  & head coach & 28 & 1,082 \\
P26   & spouse & 25 & 25 \\
P749  & parent organization & 17 & 17 \\
P793  & significant event & 16 & 600 \\
P551  & residence & 14 & 14 \\
P6    & head of government & 12 & 467 \\
P355  & child organization & 12 & 12 \\
P161  & cast member & 10 & 10 \\
P488  & chairperson & 9 & 356 \\
P69   & educated at & 4 & 165 \\
P463  & member of & 4 & 167 \\
P527  & has part(s) & 3 & 123 \\
P1037 & director / manager & 3 & 124 \\
\midrule
\multicolumn{2}{@{}l}{\textbf{Total}} & \textbf{1,000} & \textbf{22,708} \\
\bottomrule
\end{tabular}
\end{table}

\subsubsection{Experiment Sampling Strategy}
\label{app:sampling}

For our main experiments, we sample 1,000 records from the full dataset. Relations with fewer than 100 records (14 relations, 431 records) are included entirely; the remaining 569 slots are allocated proportionally among the 10 larger relations. Within each large relation, we apply time-stratified sampling (40\% newest, 30\% middle, 30\% oldest by \texttt{after\_object} start date) to ensure temporal diversity. We filter out records with empty labels or identical before/after objects. All sampling uses seed\,=\,42. Table~\ref{tab:sampled_dist} shows the resulting distribution.

\subsubsection{Data Examples}
Below we show a \textsc{TimeCF}-Factual record and its \textsc{TimeCF}-Fictional counterpart. The Fictional variant is obtained by replacing \texttt{after\_object} with a random entity from the same-relation pool while keeping all other fields unchanged.

\begin{tcolorbox}[breakable, colback=lightgray!20, colframe=darkgray!80, title={\centering \textsc{TimeCF}-Factual Example}]
\textbf{id:} 341 \\[1pt]
\textbf{relation:} P6 (head of government) \\[1pt]
\textbf{subject:} Toronto (Q172) \\[1pt]
\textbf{before\_object:} John Tory \\
\quad time\_range: 2014-12-01 to 2023-02-15 \\[1pt]
\textbf{after\_object:} Olivia Chow \\
\quad time\_range: 2023-07-12 to ...
\end{tcolorbox}

\begin{tcolorbox}[breakable, colback=cyan!8, colframe=cyan!80, title={\centering \textsc{TimeCF}-Fictional Example}]
\textbf{id:} 341 \\[1pt]
\textbf{relation:} P6 (head of government) \\[1pt]
\textbf{subject:} Toronto (Q172) \\[1pt]
\textbf{before\_object:} John Tory \\
\quad time\_range: 2014-12-01 to 2023-02-15 \\[1pt]
\textbf{after\_object:} Thomas M\"{u}ller \\
\quad time\_range: 2023-07-12 to ...
\end{tcolorbox}

\subsubsection{Comparison with Existing Datasets}
\label{app:comparison}

Table~\ref{tab:comparison} compares \textsc{TimeConflict} with existing knowledge editing datasets. Our dataset is the first to provide day-level temporal granularity with both factual and counterfactual variants.

\begin{table}[h]
\centering
\caption{Comparison with existing knowledge editing datasets.}
\label{tab:comparison}
\footnotesize
\setlength{\tabcolsep}{1.5pt}
\resizebox{\columnwidth}{!}{%
\begin{tabular}{@{}lcrccc@{}}
\toprule
\textbf{Dataset} & \textbf{Source} & \textbf{\#Rec.} & \textbf{\#Rel.} & \textbf{Granul.} & \textbf{CF} \\
\midrule
CounterFact \cite{meng2022locating} & Wikidata & 21,919 & 32 & None & \checkmark \\
MQuAKE \cite{zhong2023mquake} & Wikidata & 3,000 & 37 & None & \checkmark \\
ConflictBank \cite{su2024textttconflictbank} & Wikidata & 553K & -- & Year & -- \\
AToKe \cite{yin2024history} & YAGO3.0 & 8,820 & 13 & Year & -- \\
\rowcolor{gray!12}
\textsc{TimeCF} (Ours) & Wikidata & 22,708 & 24 & \textbf{Day} & \checkmark \\
\bottomrule
\end{tabular}%
}
\end{table}

\subsubsection{Dataset Audit and Manual Inspection}
\label{app:audit}

Beyond the hard constraints enforced during construction, we run a deterministic audit script over the \emph{entire} dataset to verify that the constraints hold at scale: temporal non-overlap holds for 99.96\% of all records, QID-level entity distinctness is 100\%, every after-object starts after the 2023-01-01 cutoff, and all records carry complete subject/before/after labels. We further verify the constraints per relation: all 24 relations independently pass (100\% QID-distinctness for every relation; non-overlap $\geq 89.8\%$ even for the smallest relations), confirming that data quality is not driven by a few head relations. The audit identified fewer than 0.1\% residual exact duplicates, which we have removed.

As a complement to the automated audit, we randomly sampled 1{,}000 records for manual inspection to ensure data validity.

\subsection{\textsc{CounterFact}$^{\dagger}$ Augmentation}
\label{app:counterfact_aug}

The original \textsc{CounterFact}~\cite{meng2022locating} is a static counterfactual knowledge-editing benchmark; we refer readers to the original paper for its full construction protocol. In this work, we use it as the basis for \textsc{CounterFact}$^{\dagger}$, a temporalized counterpart that keeps the original subject--relation--object rewrite semantics but adds an explicit historical/current distinction. This lets us test whether an editing method can introduce temporal conditioning into facts that are not temporal in the original benchmark.

\subsubsection{Augmentation Protocol}

\paragraph{Relation filtering.}
We first restrict the original relations to a subset of \textbf{12 time-sensitive relations} (e.g., occupation, employer, head of state, head of government, position held), keeping records where assigning the old and new answers to different time periods is semantically meaningful. This yields \textbf{6{,}219 records} from the original 21{,}919.

\paragraph{Temporal augmentation.}
We augment each filtered \textsc{CounterFact} record into a two-period contrast. The original answer is preserved in historical contexts, while the counterfactual target is assigned to current contexts. This turns a static subject--relation rewrite into a temporal editing case: a successful method must produce different answers for the same subject--relation pair depending on the temporal context. To reduce reliance on any single wording, we instantiate the historical/current contrast with several semantically equivalent temporal expressions (e.g., \textit{Since}/\textit{Before}, \textit{From\,$\ldots$\,onward}/\textit{Prior to}, \textit{As of}/\textit{Up to}).

\subsubsection{Data Example}

Below we show an original \textsc{CounterFact} record and its \textsc{CounterFact}$^{\dagger}$ counterpart. The augmented variant converts the static rewrite into a two-period temporal contrast.

\begin{tcolorbox}[colback=lightgray!20, colframe=darkgray!80, title={\centering Original \textsc{CounterFact} Example}]
\textbf{id:} 18329 \\[1pt]
\textbf{relation:} occupation \\[1pt]
\textbf{subject:} William Hardy Wilson \\[1pt]
\textbf{true\_object:} architect \\[1pt]
\textbf{counterfactual\_target:} journalist \\[1pt]
\textbf{edit:} architect $\rightarrow$ journalist
\end{tcolorbox}

\begin{tcolorbox}[colback=cyan!8, colframe=cyan!80, title={\centering Augmented \textsc{CounterFact} Example}]
\textbf{id:} 18329 \\[1pt]
\textbf{relation:} occupation \\[1pt]
\textbf{subject:} William Hardy Wilson \\[1pt]
\textbf{before\_object:} architect \\
\quad time\_context: ``Prior to 2027, the occupation of William Hardy Wilson was \underline{\hspace{0.8cm}}'' \\[1pt]
\textbf{after\_object:} journalist \\
\quad time\_context: ``From 2027 onward, the occupation of William Hardy Wilson has been \underline{\hspace{0.8cm}}''
\end{tcolorbox}

\subsubsection{Scope and Complementarity}

\textsc{CounterFact}$^{\dagger}$ is a \emph{synthetic} temporalization of \textsc{CounterFact}~\cite{meng2022locating}, the most widely adopted benchmark in the knowledge editing literature. By extending it to the temporal setting, we can evaluate our method on a well-established and broadly recognized data source, strengthening the comparability and credibility of our results. It complements \textsc{TimeCF}: \textsc{TimeCF} evaluates naturally occurring temporal changes, while \textsc{CounterFact}$^{\dagger}$ tests whether a method can impose temporal distinctions on originally static facts. For our main experiments, we sample 1{,}000 records from \textsc{CounterFact}$^{\dagger}$.

\begin{table*}[t]
\centering
\small
\begin{tabular}{ll|cccccc}
\toprule
\textbf{Dataset} & \textbf{Method} & \textbf{CES}$\uparrow$ & \textbf{CES-P}$\uparrow$ & \textbf{CRS}$\uparrow$ & \textbf{HES}$\uparrow$ & \textbf{HES-P}$\uparrow$ & \textbf{TC}$\uparrow$ \\
\midrule
& Pre-Edit & \meanstd{0.8}{0.3} & \meanstd{0.9}{0.2} & \meanstd{1.0}{0.3} & \meanstd{28.0}{1.4} & \meanstd{25.1}{1.1} & \meanstd{0.3}{0.2} \\
\cmidrule(l){2-8}
& Fine-Tuning & \meanstd{21.3}{1.3} & \meanstd{6.0}{0.4} & \meanstd{1.7}{0.4} & \meanstd{57.9}{1.6} & \meanstd{35.2}{1.0} & \meanstd{14.5}{1.1} \\
& ROME & \meanstd{5.7}{0.7} & \meanstd{4.3}{0.4} & \meanstd{3.5}{0.6} & \meanstd{91.8}{0.9} & \meanstd{80.9}{0.9} & \meanstd{4.2}{0.6} \\
\textsc{CounterFact}$^{\dagger}$ & MEMIT & \meanstd{80.5}{1.3} & \meanstd{71.4}{1.2} & \meanstd{43.4}{1.6} & \meanstd{81.0}{1.2} & \meanstd{72.8}{1.1} & \meanstd{63.4}{1.5} \\
& AlphaEdit & \meanstdb{99.9}{0.1} & \meanstdb{90.6}{0.5} & \meanstdu{43.0}{1.6} & \meanstdb{99.9}{0.1} & \meanstdb{91.2}{0.5} & \meanstdb{99.8}{0.1} \\
& METO & \meanstd{82.5}{1.2} & \meanstd{72.8}{1.4} & \meanstd{43.0}{1.6} & \meanstd{84.4}{1.1} & \meanstd{75.4}{1.4} & \meanstd{68.0}{1.5} \\
& \textbf{PRISM Edit} & \meanstdu{99.3}{0.3} & \meanstdu{86.9}{0.6} & \meanstdb{88.2}{1.0} & \meanstdu{99.7}{0.2} & \meanstdu{86.0}{0.6} & \meanstdu{99.0}{0.3} \\
\midrule
& Pre-Edit & \meanstd{5.1}{0.7} & \meanstd{4.4}{0.6} & \meanstd{7.0}{0.8} & \meanstd{7.4}{0.8} & \meanstd{4.4}{0.6} & \meanstd{1.3}{0.4} \\
\cmidrule(l){2-8}
& Fine-Tuning & \meanstd{3.6}{0.6} & \meanstd{2.1}{0.5} & \meanstd{3.8}{0.6} & \meanstd{4.0}{0.6} & \meanstd{4.2}{0.6} & \meanstd{0.4}{0.2} \\
& ROME & \meanstd{6.2}{0.8} & \meanstd{5.7}{0.7} & \meanstd{6.5}{0.8} & \meanstd{24.9}{1.4} & \meanstd{22.6}{1.3} & \meanstd{2.7}{0.5} \\
\textsc{TimeCF-Factual} & MEMIT & \meanstd{51.2}{1.6} & \meanstd{29.7}{1.4} & \meanstd{31.5}{1.5} & \meanstd{52.6}{1.6} & \meanstd{31.5}{1.5} & \meanstd{27.6}{1.4} \\
& AlphaEdit & \meanstdu{96.8}{0.6} & \meanstdu{62.8}{1.5} & \meanstdu{48.0}{1.6} & \meanstdu{98.1}{0.4} & \meanstdu{67.4}{1.5} & \meanstdu{95.3}{0.7} \\
& METO & \meanstd{48.7}{1.6} & \meanstd{28.7}{1.4} & \meanstd{17.0}{1.2} & \meanstd{52.4}{1.6} & \meanstd{32.6}{1.5} & \meanstd{28.1}{1.4} \\
& \textbf{PRISM Edit} & \meanstdb{99.1}{0.3} & \meanstdb{92.3}{0.8} & \meanstdb{91.2}{0.9} & \meanstdb{98.6}{0.4} & \meanstdb{91.4}{0.9} & \meanstdb{97.7}{0.5} \\
\midrule
& Pre-Edit & \meanstd{1.2}{0.3} & \meanstd{0.8}{0.3} & \meanstd{2.2}{0.5} & \meanstd{7.4}{0.8} & \meanstd{4.4}{0.6} & \meanstd{0.2}{0.1} \\
\cmidrule(l){2-8}
& Fine-Tuning & \meanstd{1.4}{0.4} & \meanstd{0.6}{0.2} & \meanstd{1.0}{0.3} & \meanstd{6.0}{0.8} & \meanstd{1.9}{0.4} & \meanstd{0.5}{0.2} \\
& ROME & \meanstd{2.1}{0.5} & \meanstd{1.2}{0.3} & \meanstd{1.8}{0.4} & \meanstd{22.8}{1.3} & \meanstd{17.9}{1.2} & \meanstd{1.1}{0.3} \\
\textsc{TimeCF-Fictional} & MEMIT & \meanstd{30.9}{1.5} & \meanstd{15.6}{1.1} & \meanstd{16.0}{1.2} & \meanstd{49.9}{1.6} & \meanstd{31.8}{1.5} & \meanstd{18.0}{1.2} \\
& AlphaEdit & \meanstdu{90.5}{0.9} & \meanstdu{52.2}{1.6} & \meanstdu{33.2}{1.5} & \meanstdb{97.6}{0.5} & \meanstdu{68.6}{1.5} & \meanstdu{88.4}{1.0} \\
& METO & \meanstd{29.0}{1.4} & \meanstd{13.9}{1.1} & \meanstd{9.4}{0.9} & \meanstd{48.7}{1.6} & \meanstd{28.6}{1.4} & \meanstd{18.8}{1.2} \\
& \textbf{PRISM Edit} & \meanstdb{98.7}{0.4} & \meanstdb{84.9}{1.1} & \meanstdb{81.8}{1.2} & \meanstdu{97.2}{0.5} & \meanstdb{86.0}{1.1} & \meanstdb{95.9}{0.6} \\
\bottomrule
\end{tabular}
\caption{Results on GPT-J (6B) for temporal knowledge editing
(1{,}000 records per dataset, sequential editing in
batches of 100). Best in \textbf{bold}, second-best
\underline{underlined}. $^{\dagger}$ denotes our temporal augmentation of
\textsc{CounterFact}~\citep{meng2022locating}.}
\label{tab:gptj}
\end{table*}

\section{Additional Experimental Results}
\label{app:other_ex}

\subsection{GPT-J Results}
\label{app:gptj}

GPT-J-6B~\cite{wang2021gptj} is a widely used benchmark model in the knowledge editing literature. We replicate the main experiment (\S\ref{sec:experiments}) on it to test cross-architecture generalization; only the editing layers are re-selected via causal tracing. As shown in Table~\ref{tab:gptj},  PRISM Edit achieves the best overall performance across multiple datasets, with particularly strong gains on CRS. On \textsc{CounterFact}$^{\dagger}$, AlphaEdit attains marginally higher CES/HES/TC (e.g., 99.8 vs.\ 99.0 on TC); with both methods above 99, these gaps likely reflect saturation rather than meaningful differences, possibly because GPT-J's relatively weak intrinsic temporal routing interacts with AlphaEdit's null-space projection in this near-saturated regime. Even here, PRISM Edit leads CRS by a wide margin (88.2 vs.\ 43.0), confirming that temporal-polysemy editing transfers across architectures.

\subsection{GPT-2 XL Results}
\label{app:gpt2xl}

GPT-2 XL (1.5B)~\cite{radford2019language} is one of the most widely used backbones in the knowledge-editing literature. We replicate the main experiment on it with the same setup, extending evaluation to a smaller and weaker model. As shown in Table~\ref{tab:gpt2xl}, PRISM Edit leads on the coexistence-critical metrics (HES, HES-P, TC) by a large margin---e.g., on \textsc{CounterFact}$^{\dagger}$ it reaches TC 84.1 and HES 96.1, versus the strongest baseline's 15.0 and 50.3. AlphaEdit attains higher CES/CES-P on \textsc{CounterFact}$^{\dagger}$, but only by collapsing the fact toward the new answer, driving its HES and TC to near zero---the very failure mode temporal editing aims to avoid.

\begin{table*}[!htbp]
\centering
\small
\begin{tabular}{ll|cccccc}
\toprule
\textbf{Dataset} & \textbf{Method} & \textbf{CES}$\uparrow$ & \textbf{CES-P}$\uparrow$ & \textbf{CRS}$\uparrow$ & \textbf{HES}$\uparrow$ & \textbf{HES-P}$\uparrow$ & \textbf{TC}$\uparrow$ \\
\midrule
& Pre-Edit & \meanstd{1.5}{0.4} & \meanstd{1.0}{0.3} & \meanstd{1.2}{0.3} & \meanstd{16.2}{1.2} & \meanstd{14.7}{1.1} & \meanstd{0.4}{0.2} \\
\cmidrule(l){2-8}
& FT & \meanstd{24.3}{1.4} & \meanstd{7.3}{0.8} & \meanstd{3.3}{0.6} & \meanstd{18.1}{1.2} & \meanstd{15.3}{1.1} & \meanstd{5.0}{0.7} \\
& ROME & \meanstd{87.4}{1.0} & \meanstd{69.8}{1.5} & \meanstd{66.8}{1.5} & \meanstd{1.8}{0.4} & \meanstd{1.8}{0.4} & \meanstd{1.0}{0.3} \\
\textsc{CounterFact}$^{\dagger}$ & MEMIT & \meanstdu{90.6}{0.9} & \meanstdu{78.0}{1.3} & \meanstdu{67.8}{1.5} & \meanstd{3.5}{0.6} & \meanstd{3.2}{0.6} & \meanstd{2.7}{0.5} \\
& AlphaEdit & \meanstdb{99.4}{0.2} & \meanstdb{87.3}{1.1} & \meanstdb{76.2}{1.3} & \meanstd{1.5}{0.4} & \meanstd{1.5}{0.4} & \meanstd{1.5}{0.4} \\
& METO & \meanstd{48.4}{1.6} & \meanstd{36.5}{1.2} & \meanstd{4.9}{0.7} & \meanstdu{50.3}{1.6} & \meanstdu{41.9}{1.2} & \meanstdu{15.0}{1.1} \\
& \textbf{PRISM Edit} & \meanstd{87.5}{1.0} & \meanstd{71.7}{1.4} & \meanstd{75.6}{1.4} & \meanstdb{96.1}{0.6} & \meanstdb{77.9}{1.3} & \meanstdb{84.1}{1.2} \\
\midrule
& Pre-Edit & \meanstd{1.7}{0.4} & \meanstd{1.2}{0.3} & \meanstd{3.0}{0.5} & \meanstd{1.9}{0.4} & \meanstd{2.1}{0.5} & \meanstd{0.2}{0.1} \\
\cmidrule(l){2-8}
& FT & \meanstd{1.7}{0.4} & \meanstd{1.2}{0.3} & \meanstd{0.6}{0.2} & \meanstd{0.7}{0.3} & \meanstd{0.4}{0.2} & \meanstd{0.0}{0.0} \\
& ROME & \meanstd{45.5}{1.6} & \meanstd{28.8}{1.4} & \meanstd{37.3}{1.5} & \meanstd{5.2}{0.7} & \meanstd{3.5}{0.6} & \meanstd{3.1}{0.5} \\
\textsc{TimeCF}-Factual & MEMIT & \meanstd{37.6}{1.5} & \meanstd{23.3}{1.3} & \meanstd{27.6}{1.4} & \meanstd{6.7}{0.8} & \meanstd{5.8}{0.7} & \meanstd{3.8}{0.6} \\
& AlphaEdit & \meanstdb{90.0}{0.9} & \meanstdu{53.3}{1.6} & \meanstdu{63.0}{1.5} & \meanstd{7.4}{0.8} & \meanstd{6.4}{0.8} & \meanstdu{6.7}{0.8} \\
& METO & \meanstd{11.4}{1.0} & \meanstd{8.8}{0.9} & \meanstd{5.9}{0.7} & \meanstdu{13.2}{1.1} & \meanstdu{8.7}{0.9} & \meanstd{3.0}{0.5} \\
& \textbf{PRISM Edit} & \meanstdu{71.6}{1.4} & \meanstdb{54.5}{1.6} & \meanstdb{70.9}{1.4} & \meanstdb{63.1}{1.5} & \meanstdb{53.5}{1.6} & \meanstdb{47.7}{1.6} \\
\midrule
& Pre-Edit & \meanstd{0.5}{0.2} & \meanstd{0.4}{0.2} & \meanstd{1.0}{0.3} & \meanstd{1.9}{0.4} & \meanstd{2.1}{0.5} & \meanstd{0.0}{0.0} \\
\cmidrule(l){2-8}
& FT & \meanstd{1.6}{0.4} & \meanstd{0.6}{0.2} & \meanstd{0.0}{0.0} & \meanstd{0.2}{0.1} & \meanstd{0.3}{0.2} & \meanstd{0.0}{0.0} \\
& ROME & \meanstd{24.0}{1.4} & \meanstd{17.8}{1.2} & \meanstd{23.7}{1.3} & \meanstd{1.6}{0.4} & \meanstd{0.7}{0.3} & \meanstd{0.3}{0.2} \\
\textsc{TimeCF}-Fictional & MEMIT & \meanstd{26.0}{1.4} & \meanstd{14.6}{1.1} & \meanstd{16.2}{1.2} & \meanstd{1.6}{0.4} & \meanstd{2.1}{0.5} & \meanstd{0.7}{0.3} \\
& AlphaEdit & \meanstdb{81.1}{1.2} & \meanstdb{45.2}{1.6} & \meanstdu{51.9}{1.6} & \meanstd{3.1}{0.5} & \meanstd{2.9}{0.5} & \meanstdu{2.8}{0.5} \\
& METO & \meanstd{5.2}{0.7} & \meanstd{4.3}{0.6} & \meanstd{3.3}{0.6} & \meanstdu{8.4}{0.9} & \meanstdu{4.7}{0.7} & \meanstd{1.0}{0.3} \\
& \textbf{PRISM Edit} & \meanstdu{58.3}{1.6} & \meanstdu{38.3}{1.5} & \meanstdb{56.9}{1.6} & \meanstdb{60.3}{1.5} & \meanstdb{48.1}{1.6} & \meanstdb{37.2}{1.5} \\
\bottomrule
\end{tabular}
\caption{Main results on GPT-2 XL (1{,}000 records per dataset, sequential editing in batches of 100). Best in \textbf{bold}, second-best \underline{underlined}. $^{\dagger}$ denotes our temporal augmentation of \textsc{CounterFact}~\citep{meng2022locating}.}
\label{tab:gpt2xl}
\end{table*}

Memory-based editors on GPT-2 XL are reported separately in Appendix~\ref{app:memory_based}.


\subsection{Qwen3-4B Results}
\label{app:qwen3}

To cover a more recent model family, we additionally evaluate Qwen3-4B-Instruct~\cite{qwen3} on all three datasets (1{,}000 records each) against Pre-Edit, MEMIT, and AlphaEdit under the standard editing setup. As shown in Tables~\ref{tab:qwen3}, PRISM Edit consistently outperforms both baselines across the reported metrics: its average TC reaches 48.7 versus AlphaEdit's 26.0 (about $1.9\times$), while MEMIT is essentially ineffective on this architecture. Unlike on GPT-2 XL, where the advantage takes the form of a current-vs-historical trade-off, PRISM Edit here also leads on CES/CES-P, so the advantage is uniform.

\begin{table*}[!htbp]
\centering
\small
\begin{tabular}{ll|cccccc}
\toprule
\textbf{Dataset} & \textbf{Method} & \textbf{CES}$\uparrow$ & \textbf{CES-P}$\uparrow$ & \textbf{CRS}$\uparrow$ & \textbf{HES}$\uparrow$ & \textbf{HES-P}$\uparrow$ & \textbf{TC}$\uparrow$ \\
\midrule
& Pre-Edit & \meanstd{1.2}{0.3} & \meanstd{0.5}{0.1} & \meanstd{0.6}{0.2} & \meanstd{30.9}{1.5} & \meanstd{28.3}{1.1} & \meanstd{0.1}{0.1} \\
\cmidrule(l){2-8}
& MEMIT & \meanstd{1.1}{0.3} & \meanstd{0.8}{0.2} & \meanstd{1.3}{0.4} & \meanstd{37.6}{1.5} & \meanstd{34.1}{1.2} & \meanstd{0.1}{0.1} \\
\textsc{CounterFact}$^{\dagger}$ & AlphaEdit & \meanstdu{74.1}{1.4} & \meanstdu{53.9}{1.1} & \meanstdu{28.3}{1.4} & \meanstdu{69.7}{1.5} & \meanstdu{56.0}{1.2} & \meanstdu{48.6}{1.6} \\
& \textbf{PRISM Edit} & \meanstdb{89.1}{1.0} & \meanstdb{72.9}{0.9} & \meanstdb{90.7}{0.9} & \meanstdb{94.2}{0.7} & \meanstdb{78.3}{0.7} & \meanstdb{84.1}{1.2} \\
\midrule
& Pre-Edit & \meanstd{8.1}{0.9} & \meanstd{7.2}{0.8} & \meanstd{4.2}{0.6} & \meanstd{6.8}{0.8} & \meanstd{6.5}{0.8} & \meanstd{1.3}{0.4} \\
\cmidrule(l){2-8}
& MEMIT & \meanstd{10.2}{1.0} & \meanstd{9.2}{0.9} & \meanstd{8.2}{0.9} & \meanstd{9.9}{0.9} & \meanstd{9.6}{0.9} & \meanstd{2.2}{0.5} \\
\textsc{TimeCF}-Factual & AlphaEdit & \meanstdu{38.7}{1.5} & \meanstdu{28.5}{1.4} & \meanstdu{22.5}{1.3} & \meanstdu{43.0}{1.6} & \meanstdu{27.3}{1.4} & \meanstdu{19.2}{1.2} \\
& \textbf{PRISM Edit} & \meanstdb{64.3}{1.5} & \meanstdb{56.4}{1.6} & \meanstdb{70.5}{1.4} & \meanstdb{59.0}{1.6} & \meanstdb{57.4}{1.6} & \meanstdb{41.2}{1.6} \\
\midrule
& Pre-Edit & \meanstd{1.7}{0.4} & \meanstd{1.3}{0.4} & \meanstd{1.0}{0.3} & \meanstd{6.8}{0.8} & \meanstd{6.5}{0.8} & \meanstd{0.4}{0.2} \\
\cmidrule(l){2-8}
& MEMIT & \meanstd{2.2}{0.5} & \meanstd{1.8}{0.4} & \meanstd{1.8}{0.4} & \meanstd{9.5}{0.9} & \meanstd{8.6}{0.9} & \meanstd{0.4}{0.2} \\
\textsc{TimeCF}-Fictional & AlphaEdit & \meanstdu{22.0}{1.3} & \meanstdu{13.4}{1.1} & \meanstdu{9.1}{0.9} & \meanstdu{38.9}{1.5} & \meanstdu{25.0}{1.4} & \meanstdu{10.2}{1.0} \\
& \textbf{PRISM Edit} & \meanstdb{43.9}{1.6} & \meanstdb{34.0}{1.5} & \meanstdb{48.2}{1.6} & \meanstdb{46.8}{1.6} & \meanstdb{48.2}{1.6} & \meanstdb{20.8}{1.3} \\
\bottomrule
\end{tabular}
\caption{Main results on Qwen3-4B-Instruct (1{,}000 records per dataset). Best in \textbf{bold}, second-best \underline{underlined}. PRISM Edit is best on every metric on all three datasets.}
\label{tab:qwen3}
\end{table*}

\subsection{Memory-Based Methods}
\label{app:memory_based}

Although PRISM Edit primarily focuses on parameter-modifying editing on the MLP pathway analyzed in \S\ref{sec:trace}, we also compare against representative memory-based editors to position our method within the broader knowledge-editing landscape. We select GRACE \cite{hartvigsen2023grace} and WISE \cite{wang2024wise} as representative baselines: GRACE attaches an external key--value codebook and retrieves a stored value when the input matches a cached key, while WISE introduces a parametric side memory together with a router that decides whether to read from the main FFN or the side memory. Since both mechanisms operate outside the MLP pathway and follow a fundamentally different design principle from parameter-modifying editors, we report them separately.

Tables~\ref{tab:memory_llama}, \ref{tab:memory_gptj}, and~\ref{tab:gpt2xl_memory} present the full comparison. Methods of this type perform reasonably on direct-evaluation metrics (CES, HES, TC), but still exhibit a clear gap on the paraphrase-robust counterparts CES-P and HES-P. We view this difference as reasonable: such approaches deliberately leave the original model parameters intact and instead allocate extra memory or auxiliary modules to host the edited knowledge, so their behavior on unseen inputs is largely determined by how well those inputs are matched against the cached entries. In contrast, PRISM Edit performs parametric editing directly on the MLP pathway analyzed in \S\ref{sec:trace}, and consistently leads on CES-P, HES-P and CRS, with the advantage being especially stable on paraphrase-robust metrics. Table~\ref{tab:memory_time} further shows that PRISM Edit also maintains a clear efficiency advantage, with a lower average per-record edit time than both GRACE and WISE.

\begin{table*}[!htbp]
\centering
\small
\begin{tabular}{ll|cccccc}
\toprule
\textbf{Dataset} & \textbf{Method} & \textbf{CES}$\uparrow$ & \textbf{CES-P}$\uparrow$ & \textbf{CRS}$\uparrow$ & \textbf{HES}$\uparrow$ & \textbf{HES-P}$\uparrow$ & \textbf{TC}$\uparrow$ \\
\midrule
& Pre-Edit & \meanstd{1.1}{0.3} & \meanstd{0.6}{0.1} & \meanstd{1.0}{0.3} & \meanstd{48.7}{1.6} & \meanstd{46.6}{1.2} & \meanstd{0.4}{0.2} \\
\cmidrule(l){2-8}
\textsc{CounterFact}$^\dagger$ & GRACE & \meanstdu{34.5}{1.5} & \meanstdu{7.3}{0.8} & \meanstd{1.0}{0.3} & \meanstdu{78.6}{1.3} & \meanstdu{52.5}{1.6} & \meanstdu{34.2}{1.5} \\
& WISE & \meanstd{3.4}{0.6} & \meanstd{2.3}{0.5} & \meanstdu{1.4}{0.4} & \meanstd{49.4}{1.6} & \meanstd{47.2}{1.6} & \meanstd{1.6}{0.4} \\
& \textbf{PRISM Edit} & \meanstdb{86.4}{1.1} & \meanstdb{75.5}{1.0} & \meanstdb{82.5}{1.2} & \meanstdb{95.5}{0.7} & \meanstdb{85.1}{0.7} & \meanstdb{82.4}{1.2} \\
\midrule
& Pre-Edit & \meanstd{9.5}{0.9} & \meanstdu{7.5}{0.8} & \meanstdu{9.2}{0.9} & \meanstd{15.7}{1.2} & \meanstdu{13.9}{1.1} & \meanstd{2.9}{0.5} \\
\cmidrule(l){2-8}
\textsc{TimeCF}-Factual & GRACE & \meanstdb{95.1}{0.7} & \meanstdu{7.5}{0.8} & \meanstdu{9.2}{0.9} & \meanstdb{96.8}{0.6} & \meanstdu{13.9}{1.1} & \meanstdb{93.5}{0.8} \\
& WISE & \meanstd{9.0}{0.9} & \meanstd{7.2}{0.8} & \meanstd{8.9}{0.9} & \meanstd{15.5}{1.1} & \meanstd{13.7}{1.1} & \meanstd{2.9}{0.5} \\
& \textbf{PRISM Edit} & \meanstdu{59.2}{1.6} & \meanstdb{48.1}{1.6} & \meanstdb{61.7}{1.5} & \meanstdu{60.8}{1.5} & \meanstdb{51.1}{1.6} & \meanstdu{37.4}{1.5} \\
\midrule
& Pre-Edit & \meanstd{1.6}{0.4} & \meanstdu{1.2}{0.3} & \meanstdu{2.0}{0.4} & \meanstd{15.7}{1.2} & \meanstdu{13.9}{1.1} & \meanstd{0.5}{0.2} \\
\cmidrule(l){2-8}
\textsc{TimeCF}-Fictional & GRACE & \meanstdb{96.7}{0.6} & \meanstdu{1.2}{0.3} & \meanstdu{2.0}{0.4} & \meanstdb{96.8}{0.6} & \meanstdu{13.9}{1.1} & \meanstdb{93.6}{0.8} \\
& WISE & \meanstd{1.6}{0.4} & \meanstdu{1.2}{0.3} & \meanstdu{2.0}{0.4} & \meanstd{15.7}{1.2} & \meanstdu{13.9}{1.1} & \meanstd{0.5}{0.2} \\
& \textbf{PRISM Edit} & \meanstdu{36.0}{1.5} & \meanstdb{25.4}{1.4} & \meanstdb{37.4}{1.5} & \meanstdu{60.2}{1.5} & \meanstdb{49.0}{1.6} & \meanstdu{22.8}{1.3} \\
\bottomrule
\end{tabular}
\caption{Memory-based editors on LLaMA-3. Best in \textbf{bold}, second-best \underline{underlined}.}
\label{tab:memory_llama}
\end{table*}

\begin{table*}[!htbp]
\centering
\small
\begin{tabular}{ll|cccccc}
\toprule
\textbf{Dataset} & \textbf{Method} & \textbf{CES}$\uparrow$ & \textbf{CES-P}$\uparrow$ & \textbf{CRS}$\uparrow$ & \textbf{HES}$\uparrow$ & \textbf{HES-P}$\uparrow$ & \textbf{TC}$\uparrow$ \\
\midrule
& Pre-Edit & \meanstd{0.8}{0.3} & \meanstd{0.9}{0.2} & \meanstd{1.0}{0.3} & \meanstd{28.0}{1.4} & \meanstd{25.1}{1.1} & \meanstd{0.3}{0.2} \\
\cmidrule(l){2-8}
\textsc{CounterFact}$^\dagger$ & GRACE & \meanstdb{100.0}{0.0} & \meanstdu{20.7}{1.3} & \meanstd{1.0}{0.3} & \meanstdb{100.0}{0.0} & \meanstd{39.5}{1.5} & \meanstdb{100.0}{0.0} \\
& WISE & \meanstd{12.7}{1.1} & \meanstd{7.7}{0.8} & \meanstdu{7.2}{0.8} & \meanstd{55.1}{1.6} & \meanstdu{47.8}{1.6} & \meanstd{4.6}{0.7} \\
& \textbf{PRISM Edit} & \meanstdu{99.3}{0.3} & \meanstdb{86.9}{0.6} & \meanstdb{88.2}{1.0} & \meanstdu{99.7}{0.2} & \meanstdb{86.0}{0.6} & \meanstdu{99.0}{0.3} \\
\midrule
& Pre-Edit & \meanstd{5.1}{0.7} & \meanstd{4.4}{0.6} & \meanstd{7.0}{0.8} & \meanstd{7.4}{0.8} & \meanstd{4.4}{0.6} & \meanstd{1.3}{0.4} \\
\cmidrule(l){2-8}
\textsc{TimeCF}-Factual & GRACE & \meanstdb{100.0}{0.0} & \meanstd{4.4}{0.6} & \meanstd{7.0}{0.8} & \meanstdb{100.0}{0.0} & \meanstd{4.4}{0.6} & \meanstdb{100.0}{0.0} \\
& WISE & \meanstd{8.2}{0.9} & \meanstdu{7.0}{0.8} & \meanstdu{7.5}{0.8} & \meanstd{13.9}{1.1} & \meanstdu{8.9}{0.9} & \meanstd{2.8}{0.5} \\
& \textbf{PRISM Edit} & \meanstdu{99.1}{0.3} & \meanstdb{92.3}{0.8} & \meanstdb{91.2}{0.9} & \meanstdu{98.6}{0.4} & \meanstdb{91.4}{0.9} & \meanstdu{97.7}{0.5} \\
\midrule
& Pre-Edit & \meanstd{1.2}{0.3} & \meanstd{0.8}{0.3} & \meanstd{2.2}{0.5} & \meanstd{7.4}{0.8} & \meanstd{4.4}{0.6} & \meanstd{0.2}{0.1} \\
\cmidrule(l){2-8}
\textsc{TimeCF}-Fictional & GRACE & \meanstdb{100.0}{0.0} & \meanstd{0.8}{0.3} & \meanstd{2.2}{0.5} & \meanstdb{100.0}{0.0} & \meanstd{4.4}{0.6} & \meanstdb{100.0}{0.0} \\
& WISE & \meanstd{3.9}{0.6} & \meanstdu{2.1}{0.5} & \meanstdu{2.5}{0.5} & \meanstd{15.5}{1.1} & \meanstdu{11.5}{1.0} & \meanstd{1.3}{0.4} \\
& \textbf{PRISM Edit} & \meanstdu{98.7}{0.4} & \meanstdb{84.9}{1.1} & \meanstdb{81.8}{1.2} & \meanstdu{97.2}{0.5} & \meanstdb{86.0}{1.1} & \meanstdu{95.9}{0.6} \\
\bottomrule
\end{tabular}
\caption{Memory-based editors on GPT-J. Best in \textbf{bold}, second-best \underline{underlined}.}
\label{tab:memory_gptj}
\end{table*}

\begin{table*}[!htbp]
\centering
\small
\begin{tabular}{ll|cccccc}
\toprule
\textbf{Dataset} & \textbf{Method} & \textbf{CES}$\uparrow$ & \textbf{CES-P}$\uparrow$ & \textbf{CRS}$\uparrow$ & \textbf{HES}$\uparrow$ & \textbf{HES-P}$\uparrow$ & \textbf{TC}$\uparrow$ \\
\midrule
& Pre-Edit & \meanstd{1.5}{0.4} & \meanstd{1.0}{0.3} & \meanstd{1.2}{0.3} & \meanstd{16.2}{1.2} & \meanstd{14.7}{1.1} & \meanstd{0.4}{0.2} \\
\cmidrule(l){2-8}
\textsc{CounterFact}$^\dagger$ & GRACE & \meanstdu{12.5}{1.0} & \meanstdu{3.2}{0.3} & \meanstd{1.2}{0.3} & \meanstdu{40.4}{1.6} & \meanstdu{19.5}{0.8} & \meanstdu{12.0}{1.0} \\
& WISE & \meanstd{5.0}{0.7} & \meanstd{2.1}{0.3} & \meanstdu{2.8}{0.5} & \meanstd{25.4}{1.4} & \meanstd{15.7}{0.8} & \meanstd{0.9}{0.3} \\
& \textbf{PRISM Edit} & \meanstdb{87.5}{1.0} & \meanstdb{71.7}{1.4} & \meanstdb{75.6}{1.4} & \meanstdb{96.1}{0.6} & \meanstdb{77.9}{1.3} & \meanstdb{84.1}{1.2} \\
\midrule
& Pre-Edit & \meanstd{1.7}{0.4} & \meanstd{1.2}{0.3} & \meanstd{3.0}{0.5} & \meanstd{1.9}{0.4} & \meanstd{2.1}{0.5} & \meanstd{0.2}{0.1} \\
\cmidrule(l){2-8}
\textsc{TimeCF}-Factual & GRACE & \meanstdb{100.0}{0.0} & \meanstd{1.2}{0.3} & \meanstd{3.0}{0.5} & \meanstdb{100.0}{0.0} & \meanstdu{2.1}{0.5} & \meanstdb{100.0}{0.0} \\
& WISE & \meanstd{1.8}{0.4} & \meanstdu{1.6}{0.4} & \meanstdu{4.6}{0.7} & \meanstd{1.8}{0.4} & \meanstdu{2.1}{0.5} & \meanstd{0.1}{0.1} \\
& \textbf{PRISM Edit} & \meanstdu{71.6}{1.4} & \meanstdb{54.5}{1.6} & \meanstdb{70.9}{1.4} & \meanstdu{63.1}{1.5} & \meanstdb{53.5}{1.6} & \meanstdu{47.7}{1.6} \\
\midrule
& Pre-Edit & \meanstd{0.5}{0.2} & \meanstd{0.4}{0.2} & \meanstd{1.0}{0.3} & \meanstd{1.9}{0.4} & \meanstd{2.1}{0.5} & \meanstd{0.0}{0.0} \\
\cmidrule(l){2-8}
\textsc{TimeCF}-Fictional & GRACE & \meanstdb{100.0}{0.0} & \meanstdu{0.4}{0.2} & \meanstd{1.0}{0.3} & \meanstdb{100.0}{0.0} & \meanstdu{2.1}{0.5} & \meanstdb{100.0}{0.0} \\
& WISE & \meanstd{0.6}{0.2} & \meanstdu{0.4}{0.2} & \meanstdu{1.4}{0.4} & \meanstd{2.1}{0.5} & \meanstd{1.8}{0.4} & \meanstd{0.0}{0.0} \\
& \textbf{PRISM Edit} & \meanstdu{58.3}{1.6} & \meanstdb{38.3}{1.5} & \meanstdb{56.9}{1.6} & \meanstdu{60.3}{1.5} & \meanstdb{48.1}{1.6} & \meanstdu{37.2}{1.5} \\
\bottomrule
\end{tabular}
\caption{Memory-based editors on GPT-2 XL. Best in \textbf{bold}, second-best \underline{underlined}.}
\label{tab:gpt2xl_memory}
\end{table*}

\begin{table}[t]
\centering
\small
\begin{tabular}{llc}
\toprule
\textbf{Model} & \textbf{Method} & \textbf{Time (s)}$\downarrow$ \\
\midrule
\textsc{LLaMA-3} & GRACE & 8.06 \\
& WISE & 25.86 \\
& \textbf{PRISM Edit} & \textbf{5.44} \\
\midrule
\textsc{GPT-J} & GRACE & 27.22 \\
& WISE & 20.87 \\
& \textbf{PRISM Edit} & \textbf{3.33} \\
\midrule
\textsc{GPT-2 XL} & GRACE & 20.88 \\
& WISE & 15.54 \\
& \textbf{PRISM Edit} & \textbf{1.37} \\
\bottomrule
\end{tabular}
\caption{Average per-record edit time for memory-based methods and PRISM Edit.}
\label{tab:memory_time}
\end{table}

\subsection{Attention-Route Severing}
\label{app:severing}

On a PRISM-edited LLaMA-3-8B model, we sever the attention route from the prediction (last) token to time-expression tokens in different layer ranges, keeping the edited weights and all other components unchanged. The experiment uses 100 randomly sampled records from temporally augmented \textsc{CounterFact}; paraphrase metrics are averaged over multiple paraphrases, so some values are non-integers.

\begin{table}[h]
\centering
\small
\resizebox{\columnwidth}{!}{%
\begin{tabular}{lcccccc}
\toprule
Severed range & CRS & CES & CES-P & HES & HES-P & TC \\
\midrule
baseline & 76.0 & 81.0 & 64.2 & 92.0 & 77.2 & 74.0 \\
L2--8  & 76.0 & 30.0 & 30.8 & 66.0 & 62.6 & 23.0 \\
L9--15 & 76.0 & 79.0 & 62.4 & 92.0 & 77.0 & 74.0 \\
L16--31 & 76.0 & 83.0 & 65.4 & 89.0 & 78.0 & 73.0 \\
\bottomrule
\end{tabular}%
}
\caption{Attention-route severing on the PRISM-edited model}
\label{tab:sever}
\end{table}

Severing the L2--8 route reduces TC from $74.0$ to $23.0$ and CES from $81.0$ to $30.0$, while CRS remains unchanged at $76.0$ (Table~\ref{tab:sever}). The stability of CRS indicates that the intervention selectively weakens outputs requiring temporal-context modulation rather than damaging the edit itself. In contrast, severing L9--15 or L16--31 has almost no effect on TC. Combined with the tracing analysis, this indicates that the necessary delivery of temporal context is completed in the early segment, while transport remains active up to L15. This evidence establishes the functional involvement of the route, complementing the tracing analysis.

\subsection{Activation-Patching Probe}
\label{app:patching}
This experiment isolates the representation-level claim from the full weight-writing pipeline. The experimental setup is identical to the severing experiment above: the same PRISM-edited LLaMA-3-8B model and the same 100 randomly sampled records from temporally augmented \textsc{CounterFact}. We directly patch the PRISM joint edit vector into the L8 subject-token hidden state and test whether the unchanged downstream network can read it out as different answers under new-time, old-time, and bare contexts. All controls use the same patch position, prompts, and decision criterion, differing only in the patched vector: \emph{new\_only} uses the same optimization but sets the old/bare weights to zero; \emph{no\_patch} performs the original forward pass; \emph{random} patches a Gaussian vector with the same norm as the joint vector.

\begin{table}[h]
\centering
\small
\resizebox{\columnwidth}{!}{%
\begin{tabular}{lcccc}
\toprule
Control & new & old & bare & All-route \\
\midrule
joint (patched at L8) & 100\% & 98\% & 100\% & 98\% \\
new\_only & 100\% & 2\% & 97\% & 1\% \\
no\_patch & 13\% & 92\% & 10\% & 4\% \\
random & 19\% & 82\% & 12\% & 2\% \\
\bottomrule
\end{tabular}%
}
\caption{Activation-patching probe: success rates by context type.}
\label{tab:patch}
\end{table}

The joint vector achieves $98\%$ all-route success while all controls are near zero ($1$--$4\%$) (Table~\ref{tab:patch}). The failure modes are informative: \emph{new\_only} succeeds on new-time and bare prompts but fails almost completely on old-time prompts, showing that injecting a new-answer representation alone is insufficient for historical preservation; the relatively high \emph{old}-success of \emph{no\_patch} and \emph{random} more likely reflects the base model's prior preference for the old answer under old-time prompts than genuine three-way routing. This probe thus supports that the joint vector is not an answer overwrite but a shared representation that can be conditionally read out by temporal context.

\subsection{Ablation on Context Templates}
\label{app:ctx_ablation}

In MEMIT~\citep{meng2023memit}, the target value vector is optimized by averaging over $P$ random prefix contexts prepended to the templated prompt, encouraging the edit to generalize beyond a single surface form. Since PRISM Edit already jointly optimizes $v^*$ over multiple temporal conditions, this multi-condition diversity may subsume the role of random prefixes. We test this hypothesis by fixing $\lambda_\text{old}{=}\lambda_\text{bare}{=}2$ and varying the number of paraphrased context templates per temporal condition from 1 to 10.

\begin{table}[t]
\centering
\small
\setlength{\tabcolsep}{4pt}
\begin{tabular}{c|cccccc}
\toprule
\#\,ctx & CES & CES-P & HES & HES-P & CRS & TC \\
\midrule
\rowcolor{gray!15} \textbf{1} & 79.0 & 55.0 & \textbf{72.0} & \textbf{54.0} & 74.0 & \textbf{59.0} \\
2 & \textbf{80.0} & \textbf{59.0} & 65.0 & 45.0 & 78.0 & 52.0 \\
3 & 77.0 & 52.0 & 68.0 & 48.0 & 79.0 & 57.0 \\
6 & 74.0 & 50.0 & 64.0 & 48.0 & 75.0 & 49.0 \\
10 & 71.0 & 45.0 & 64.0 & 43.0 & \textbf{80.0} & 49.0 \\
\bottomrule
\end{tabular}

\caption{Context-template ablation on \textsc{TimeConflict}-Factual ($N{=}100$, \%). Shaded row is the default.}
\label{tab:ablation_ctx}
\end{table}
 
Results confirm this hypothesis. Increasing templates from 1 to 2 marginally improves CES/CES-P, yet degrades historical retention (HES/HES-P) and overall TC. At 10 templates, CRS peaks but TC drops by 10 points. The single-template default achieves the best overall balance (72.0/54.0/59.0 on HES/HES-P/TC), indicating that PRISM Edit's multi-condition objective already provides sufficient optimization diversity---additional prefix augmentation only dilutes the temporal signal at the write site.

\subsection{Success and Failure Analysis}
\label{app:failure_analysis}

To understand when PRISM Edit fails, we analyze the same fixed set of 100 randomly sampled \textsc{CounterFact}$^{\dagger}$ records used by our other diagnostics (LLaMA-3-8B). On this subset PRISM Edit attains a TC of 74.0\%: 74 records succeed and 26 fail. We ask whether failures arise because the learned polysemous vector $\boldsymbol{v}^{*}$ itself cannot be routed by temporal context, or because the weight-writing step fails to deploy an otherwise routeable vector. To this end, we evaluate each failed record at two levels: the \emph{representation level}, where the learned joint vector is patched directly into the subject-token hidden state (the activation-patching protocol of \S\ref{app:patching}) and the unchanged downstream network is tested for correct routing; and the \emph{deployment level}, i.e., the actual weight-edited model.

\begin{table}[h]
\centering
\small
\begin{tabular}{lcc}
\toprule
\textbf{Failure type} & \textbf{\#Cases} & \textbf{\% of failures} \\
\midrule
Representation-level & 1 & 3.8 \\
Deployment-level & 25 & 96.2 \\
\bottomrule
\end{tabular}
\caption{Failure diagnosis of the 26 TC-failed records. \emph{Representation-level}: $\boldsymbol{v}^{*}$ is not routeable even under direct activation patching; \emph{deployment-level}: $\boldsymbol{v}^{*}$ remains routeable under patching but the deployed weight-edited model fails TC. Most failures are not caused by the learned vector lacking temporal routeability.}
\label{tab:failure_type}
\end{table}

\begin{table}[h]
\centering
\small
\begin{tabular}{lccc}
\toprule
\textbf{Group} & \textbf{$n$} & \textbf{Base sens.} & \textbf{All-route} \\
\midrule
TC success & 74 & 1.143 & 98.6 \\
TC failure & 26 & 0.810 & 96.2 \\
\bottomrule
\end{tabular}
\caption{Group-level comparison of successful vs.\ failed records. \emph{Base sens.} is the temporal sensitivity of the unedited base model: $\Delta_{\text{cur}}-\Delta_{\text{hist}}$, where $\Delta_{\text{cur}}=\log p(\text{new}\,|\,\text{current-time prompt})-\log p(\text{old}\,|\,\text{current-time prompt})$ and $\Delta_{\text{hist}}$ is defined analogously under the historical-time prompt. \emph{All-route} is the fraction of records whose joint vector is routed correctly under all temporal contexts in the activation-patching probe.}
\label{tab:failure_group}
\end{table}

As shown in Tables~\ref{tab:failure_type} and~\ref{tab:failure_group}, most failures are not representation-level: under direct patching, $\boldsymbol{v}^{*}$ remains routeable on 96.2\% of the TC-failed records. A more plausible locus of failure is the transition from activation patching to actual weight writing---closed-form write approximation, interference among the 100 simultaneous edits, or insufficient post-write separation between the new- and old-time readouts. Within the failure group, the degradation is also asymmetric across temporal directions: CES drops to 26.9\% while HES remains 69.2\%, i.e., deployed failures predominantly lose the new answer rather than the old one. We stress that this is a general property of locate-then-edit weight writing rather than a PRISM-specific weakness: under the same writing mechanism, PRISM Edit still substantially outperforms prior editors (Table~\ref{tab:main}). Failed cases do exhibit lower base temporal sensitivity than successful ones (0.810 vs.\ 1.143), indicating that weak base temporal modulation raises failure risk; however, it does not determine failure, since most failed cases remain routeable at the activation level.

\section{PRISM Edit Algorithm}
\label{app:prism_algorithm}
\begin{algorithm}[H]
\caption{PRISM Edit}
\label{alg:prism}
\small
\linespread{1.2}\selectfont
\begin{flushleft}
\textbf{Input:} edits $\mathcal{E}{=}\{e_b\}_{b=1}^{B}$ with $e_b{=}(s_b,r_b,\{(t_{b,i},o_{b,i},\lambda_{b,i})\}_{i=1}^{N_b})$; pretrained model $\theta$; target layer $\ell^\star$; layers to edit $\mathcal{L}$; null-space projection statistics $\{C_\ell,P_\ell\}_{\ell\in\mathcal{L}}$\\
\textbf{Output:} edited model $\theta'$
\end{flushleft}
\begin{algorithmic}[1]
\For{$e_b\in\mathcal{E}$}
  \State $\mathcal{D}_b\!\leftarrow\!\{(p_i(s_b,r_b,t_{b,i}),o_{b,i},\lambda_{b,i})\}_{i=1}^{N_b}$
  \State $\boldsymbol{v}_{0,b}\!\leftarrow\!h_{\ell^\star}(p_{\text{bare},b},\mathrm{subj\_last})$;\; $\delta_b\!\leftarrow\!\mathbf{0}$
  \State $\delta_b^*\!\leftarrow\!\arg\min_{\delta_b}\!\sum_{(p,o,\lambda)\in\mathcal{D}_b}\!\lambda\,\mathcal{L}(p,o;\delta_b){+}\Omega(\delta_b)$
  \Statex \quad where $\mathcal{L}(p,o;\delta_b){=}{-}\log\mathbb{P}_{G(\boldsymbol{v}_{0,b}+\delta_b)}(o\mid p)$
  \State $\boldsymbol{v}_b^*\leftarrow \boldsymbol{v}_{0,b}+\delta_b^*$
\EndFor
\For{$\ell_w \in \mathcal{L}$}
  \For{$e_b\in\mathcal{E}$}
    \State $h_{b,\ell_w}^{\mathrm{cur}}\!\leftarrow\!h_{\ell_w}(p_{\text{bare},b},\mathrm{subj\_last})$
    \State $k_{b,\ell_w}\!\leftarrow\!\frac{1}{P}\sum_{q=1}^{P}k_{\ell_w}(x_q\!\oplus\!s_b)$
    \State $r_{b,\ell_w}\!\leftarrow\!(\boldsymbol{v}_b^*{-}h_{b,\ell_w}^{\mathrm{cur}})/|\{\ell{\in}\mathcal{L}{:}\ell{\ge}\ell_w\}|$
  \EndFor
  \State $K_{\ell_w}\!\leftarrow\![k_{1,\ell_w},\!\ldots\!,k_{B,\ell_w}]$
  \State $R_{\ell_w}\!\leftarrow\![r_{1,\ell_w},\!\ldots\!,r_{B,\ell_w}]$
  \State $\Delta W_{\ell_w}\!\leftarrow\!R_{\ell_w}K_{\ell_w}^{\top}P_{\ell_w}M_{\ell_w}^{-1}$
  \Statex \quad where $M_{\ell_w}{=}P_{\ell_w}(C_{\ell_w}{+}K_{\ell_w}K_{\ell_w}^{\top}){+}\eta I$
  \State $W_{\ell_w}\leftarrow W_{\ell_w}+\Delta W_{\ell_w}$
\EndFor
\State $\theta'\leftarrow\theta$ with updated $\{W_{\ell_w}\}_{\ell_w\in\mathcal{L}}$
\State \Return edited model $\theta'$
\end{algorithmic}
\end{algorithm}

\noindent\textbf{Notation.}~ $h_\ell(p,t)\in\mathbb{R}^d$ denotes the hidden state at layer $\ell$ and token position $t$ for input $p$. The \emph{target layer} $\ell^\star$ is where the polysemous value $\boldsymbol{v}^\ast$ is optimized, and $\mathcal{L}$ is the set of \emph{layers to edit}. For edit $b$, $p_{\text{bare},b}$ is the bare prompt built from $(s_b, r_b)$ without temporal tokens, $\mathrm{subj\_last}$ marks the last-subject-token position, and $\lambda_{b,i}$ weights the $i$-th target pair $(t_{b,i}, o_{b,i})$. The key $k_\ell(\cdot)$ is the MLP key at layer $\ell$, averaged over random context templates $\{x_q\}_{q=1}^{P}$. Finally, $C_\ell$ and $P_\ell$ are the pre-computed key covariance and null-space projector at layer $\ell$, and $\eta$ is the L2 regularization coefficient.

\end{document}